%% file: acl_latex.tex
\pdfoutput=1
\documentclass[11pt]{article}

\usepackage{acl}

\usepackage{times}
\usepackage{latexsym}
\usepackage{graphicx}
\usepackage{amsmath}
\usepackage{multirow}
\usepackage{hyperref}
\usepackage{xcolor} 
\usepackage{booktabs}
\usepackage{algorithm}
\usepackage{algpseudocode}
\usepackage{adjustbox}
\usepackage{geometry}
\usepackage{lmodern}
\usepackage{xcolor}

\usepackage[T2A]{fontenc} % Required for Cyrillic fonts
\usepackage[russian, english]{babel} % Language support
\usepackage{CJKutf8} % Package for CJK languages

\usepackage{rotating}   % Enables 90-degree rotated tables
\usepackage{geometry}   % Adjusts page margins for fitting
\usepackage{array}      % Improves column formatting
\usepackage{tabularx}   % Allows for flexible column widths

\usepackage[T1]{fontenc}
\usepackage{tempora}
\usepackage[utf8]{inputenc}

\usepackage{microtype}

\usepackage{inconsolata}

\usepackage{graphicx}

\title{Deep Temporal Reasoning in Video Language Models: A Cross-Linguistic Evaluation of Action Duration and Completion through Perfect Times}

\author{Olga Loginova \\
  University of Trento \\
  \texttt{olga.loginova@unitn.it} \\\AND
  Sofía Ortega Loguinova \\
  Maastricht University \\
  \texttt{s.ortegaloguinova@student.maastrichtuniversity.nl} \\}

\begin{document}
\maketitle

\begin{abstract}
Human perception of events is intrinsically tied to distinguishing between completed (perfect and telic) and ongoing (durative) actions, a process mediated by both linguistic structure and visual cues. In this work, we introduce the \textbf{Perfect Times} dataset, a novel, quadrilingual (English, Italian, Russian, and Japanese) multiple-choice question-answering benchmark designed to assess video-language models (VLMs) on temporal reasoning. By pairing everyday activity videos with event completion labels and perfectivity-tailored distractors, our dataset probes whether models truly comprehend temporal dynamics or merely latch onto superficial markers. Experimental results indicate that state-of-the-art models, despite their success on text-based tasks, struggle to mirror human-like temporal and causal reasoning grounded in video. This study underscores the necessity of integrating deep multimodal cues to capture the nuances of action duration and completion within temporal and causal video dynamics, setting a new standard for evaluating and advancing temporal reasoning in VLMs.
\end{abstract}

\input{introduction}
\input{related_work}
\input{method}

\input{results}
\input{conclusion}
\input{limitations}

\section{Ethics Statement}
\label{sec:ethics}

We  introduce a new benchmark dataset derived from the video datasets \textit{Charades} and \textit{Action Genome}. We re-annotate their videos to support the new evaluation of both open- and closed-source vision-language models (VLMs). In doing so, we strictly adhere to the licensing terms of the original datasets, ensuring that our derivative work complies with all copyright and usage restrictions.

Our annotation process involved trained annotators following the guidelines aimed at ensuring consistency. Despite these efforts, we acknowledge that any human annotation process may introduce subjective interpretations. We therefore encourage users of our dataset to consider potential annotation biases when interpreting experimental results.

The evaluation of VLMs, particularly those that generate open-ended text, carries inherent risks. We have taken measures to mitigate such risks by prompting and conducting evaluations. We promote transparency through the public release of our dataset and code. Our intention is to foster reproducible research and to provide a resource that can contribute to improving the trustworthiness and robustness of VLMs.

\section*{Acknowledgments}

Olga Loginova thanks Amazon Alexa for supporting her research through a generous donation to Raffaella Bernardi. The authors thank the native speakers of the target languages, both annotators and consultants with relevant backgrounds: Angelo Noè, Vasili Noè, Elizaveta Loginova, Maya Udaka, and Cristina Crippa. We are especially grateful to Angelo Noè for his considerable support throughout this project.

% Bibliography entries for the entire Anthology, followed by custom entries
%\bibliography{anthology,custom}
% Custom bibliography entries only
\bibliography{custom}

\appendix
\input{latex/appendix}

\end{document}

%% file: introduction.tex
\section{Introduction}
\label{sec:intro}

Understanding how events unfold in time requires a detailed analysis of their both sequential and causal relationships. Sequential events are not simply arranged chronologically; rather, one event often triggers the next upon reaching its completion. Moens and Steedman~\cite{moens-steedman-1988-temporal} highlighted that human memory organizes actions based on contingency, their cause-effect relationships, where a cause reaches its culmination before triggering an effect. This causal linkage is encoded in language through grammatical time and, critically, through aspect, specifically, perfectivity and its semantic correlate, telicity. Telicity refers to a property of verb phrases that denotes a definitive endpoint (e.g., ``to put something somewhere"), while atelic (or durative) expressions (e.g., ``to hold something") lack such clear termination. This property is deeply woven into the language~\cite{Tenny1994AspectualRA}.

Temporal relations, therefore, are not conveyed solely through grammatical time; they are also robustly signaled by aspect. Yet, language often omits critical information that visual modality can provide. Observing how events unfold in time, like in videos, offers dynamic cues, such as key frame transitions and motion patterns, that decisively indicate whether an action has been completed or is still ongoing. This multimodal integration is indispensable for resolving ambiguities inherent in linguistic descriptions of events.

\begin{figure}[htbp]
\centering
\includegraphics[width=\columnwidth]{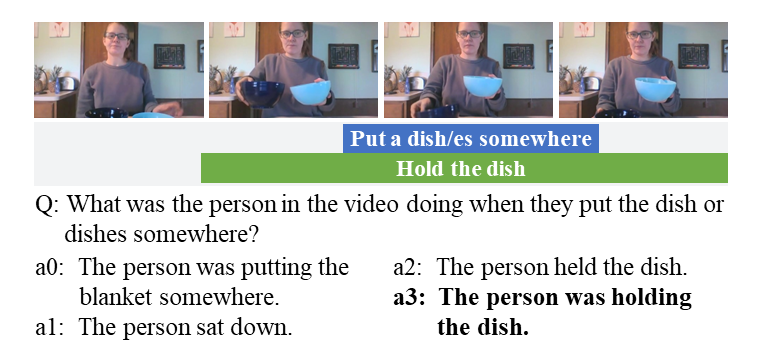}
\caption{An example from the Perfect Times dataset illustrating the interaction between a completed action (\textit{to put a dish or dishes somewhere}) and a durative action (\textit{to hold the dish}). The blue and green stripes indicate temporal progression. The correct answer highlighted in bold.}
\label{fig:example}
\end{figure}

To evaluate these phenomena, we introduce the \textbf{Perfect Times} dataset, a multiple-choice question-answering (MCQA) benchmark designed specifically for video-language models (VLMs). Our questions are constructed as complex sentences that juxtapose two actions, thereby probing the interplay between 
 verb forms and temporal conjunctions in main and dependent clauses. Twelve carefully designed templates systematically cover all possible temporal relations between the two actions in accordance with the universal Allen's interval algebra ~\cite{allen1984towards}. Moreover, our study adopts a quadrilingual approach (English, Italian, Russian, and Japanese) to capture the diverse ways in which languages encode time and aspect. For instance, Russian frequently encodes perfectivity at the lexical level, making markers of completion predefined in verbs, while Italian involves a sophisticated interaction between time, mood, and aspect, as it has more branched \textit{concordanza dei tempi} (the sequence of tenses) than English. In Japanese, the commonplace understated and ambiguous linguistic encoding necessitates stronger reliance on visual context to disambiguate temporal relations.

By combining visual data with linguistically complex questions, our work addresses the following questions:
\begin{itemize}
    \item How do VLMs leverage linguistic markers and visual cues to distinguish sequential from simultaneous actions?
    \item Are there cross-linguistic differences in the interpretation of temporal relations based on grammatical encoding of perfectivity?
    \item How closely do model predictions align with human understanding of event completion expressed in grammatical time and aspect?
\end{itemize}

Figure ~\ref{fig:example} presents an example of an English question-answer (QA) pair from the Perfect Times dataset. % \footnote{Both the dataset generation code and the dataset itself are published on \url{https://github.com/ologin/Perfect-Time.}}
In addition to determining the correct action to answer the question in the video, the model must catch the matching forms of semantic or grammatical aspect in the question and the answer option. For human native speakers, making such comparisons is not difficult, but none of the tested state-of-the-art models reach the 50\% accuracy threshold.

Our comprehensive, cross-linguistic approach thus aims to set a new benchmark for evaluating multimodal temporal reasoning in VLMs\footnote{The dataset and codes are available at https://github.com/ologin/PerfectTimes}.

%% file: related_work.tex
\section{Related Works}
\label{related}

\subsection{Cognitive Approach}
\label{sec:cog}

Several studies use the visual world paradigm to examine how aspect influences real-time language comprehension. \citet{Foppolo2021DrawAS} show that visual and linguistic signals jointly shape the interpretation of completed actions, guiding anticipatory eye movements through verb aspect and visual cues. Similarly, \citet{Foppolo2016TheIP} find that Italian adults rapidly focus on images of finished events upon hearing the corresponding verb. In an eye-tracking study, \citet{bosch_chailleux_foppolo_2021} report that while Italian children use verb semantics and aspect to anticipate outcomes, their processing of aspectual information lags behind basic lexical semantics. \citet{vanHout} challenges uniformity in aspect acquisition by demonstrating that Italian children acquire perfectivity later than their Polish and Dutch peers. \citet{Minor2022TemporalIA} further show that perfective and imperfective aspects influence listeners' expectations differently across Russian, Spanish, and English, with Russian perfectives strongly indicating completion and the English simple past being less reliable. Finally, \citet{Chang2023VisualHF} provide evidence from animation experiments that visual cues of goal information affect the choice of past versus progressive verb forms in Japanese. These findings contributing to cognitive science  illustrate that VLMs may have greater potential to replicate human-like language processing than text-only large language models (LLMs).

\subsection{Time and Aspect in Transformers}

\citet{zhao-etal-2021-pretrained} compared aspect interpretation in humans and transformers using verb tenses and resultative structures. Humans responded to all cues, while transformers were sensitive to explicit telicity but struggled with resultatives. \citet{lombardi2023agentivita} found that the transformer-based \cite{vaswani2023attention} Italian model GilBERTo performed similarly to humans on inherently telic and atelic verbs, yet had difficulty with context-dependent verbs, suggesting its limitations in handling nuanced larger temporal contexts. \citet{metheniti-etal-2022-time} showed that transformer-based models classify activities by duration and perfectivity in English and French with over 80\% accuracy.

However, all these studies are on text-only LLMs, and the experiments lack a dedicated multimodal benchmark that fully captures all possible temporal relations grounded in the physical world.

\begin{table*}
    \centering
    \scriptsize
    \caption{Examples of questions and answers in Perfect Times generated by temporal and aspectual templates with respect to the telicity markers (t: telic, a: atelic). MCA refers to the action in the main clause, and DCA corresponds to the action in the dependent clause.}
    \label{tab:temporal_aspectual_examples}
    \renewcommand{\arraystretch}{1.2}
    \setlength{\tabcolsep}{4pt}
    \begin{tabular}{ccc p{8.5cm} p{5cm}}
        \hline
        \textbf{Template} & \textbf{MCA} & \textbf{DCA} & \textbf{Question} & \textbf{Answer} \\
        \hline
        \multicolumn{5}{c}{\textbf{Precedence}} \\
        \hline
        3 & t & a & What had the person in the video done before holding the sandwich? & The person had opened the refrigerator. \\
        4 & t & t & What had the person in the video done before the other person walked through a doorway? & The person had closed the laptop. \\
        6 & a & a & What had the person in the video been doing before playing with a phone or camera? & The person had been watching the television. \\
        7 & a & t & What had the person in the video been doing before turning off the light? & The person had been playing with a phone or camera. \\
        \hline
        \multicolumn{5}{c}{\textbf{Succession}} \\
        \hline
        1 & t & a & What did the person in the video do after tidying up the table? & The person put the food somewhere. \\
        2 & t & t & What did the person in the video do after they had opened the box? & The person put the bag somewhere. \\
        10 & a & a & What was the person in the video doing after sitting in a chair? & The person was sitting on the floor. \\
        11 & a & t & What was the person in the video doing after taking the cup from somewhere? & The person was drinking from the cup. \\
        9 & a & t & What was the person in the video doing when the other person closed the door? & The person was watching the television. \\
        \hline
        \multicolumn{5}{c}{\textbf{Simultaneity}} \\
        \hline
        5 & t & a & What did the person in the video do while the other person was sitting on the floor? & The person opened the door. \\
        8 & a & a & What was the person in the video doing while holding the book? & The person was holding the bag. \\
        12 & t & t & What did the person in the video do when they grasped onto a doorknob? & The person opened the door. \\
        \hline
    \end{tabular}
\end{table*}

\subsection{Action Recognition and MCQA for Video}

Numerous datasets offer short video clips for action recognition \cite{Kay2017TheKH,Liu2021FineActionAF,Heilbron2015ActivityNetAL,Damen2020TheED,Sigurdsson2016HollywoodIH}. Kinetics \cite{Kay2017TheKH} provides general action recognition (~10-second clips); FineAction \cite{Liu2021FineActionAF} and ActivityNet \cite{Heilbron2015ActivityNetAL} deliver fine-grained temporal annotations. EPIC-Kitchens \cite{Damen2020TheED} and Charades \cite{Sigurdsson2016HollywoodIH} target specific scenarios for classification and localization.

These datasets lay the foundation for the video MCQA datasets. Templated questions appear in MSVD-QA and MSRVTT-QA \cite{Xu2017VideoQA}, STVQ \cite{Jang2019VideoQA}, and STAR \cite{Wu2021STARAB}, while manually annotated formats are used in ActivityNet-QA \cite{Yu2019ActivityNetQAAD}, TVQA \cite{Lei2018TVQALC}, NExT-QA \cite{Xiao2021NExTQANP}, Charades-SRL-QA \cite{Sadhu2021VideoQA}, and Action Genome \cite{Ji2019ActionGA}. As causal and temporal questions constitute only a subset of these benchmarks, temporal relations are often limited to simple sequential "before/after" or generic "when" questions. This "before/after" focus is also prevalent in video instruction datasets, which mainly assess the understanding of instructional step order and restrict temporal relations to sequence by design (HowTo1Million \cite{miech19howto100m}, InstructionBench \cite{Wei2025InstructionBenchAI}, YouCookQA \cite{Wang2021HowTM}).

Causal and temporal questions are frequently embedded within broader datasets aimed at evaluating diverse reasoning skills. Such datasets include Video-MME \cite{fu2024videommefirstevercomprehensiveevaluation} and Perception Test \cite{patraucean2023perception}, which span a variety of reasoning types, as well as TVBench \cite{Cores2024LostIT}, TemporalBench \cite{cai2024temporalbench}, and TempCompass \cite{liu2024tempcompassvideollmsreally}, which assess models’ grasp of temporal dynamics through complex, though not exclusively temporal, questions.

Moreover, most existing datasets are primarily in English and do not adequately address temporal dynamics from a linguistic perspective.

Our multilingual dataset addresses these gaps. It consists of Charades videos annotated with Action Genome action classes and applies STAR-inspired temporal templates that systematically incorporate aspect, a crucial linguistic dimension previously overlooked. Grounded in Allen’s interval algebra and informed by linguistic theory, Perfect Times is the first to systematically align temporal video dynamics with linguistic aspect.

\subsection{Video Language Models}

Video Language Models integrate visual and textual processing through three key components: a pre-trained visual encoder, a pre-trained large language model (LLM), and a modality interface \cite{Zhong2022VideoQA}. The visual encoder \cite{Radford2021LearningTV,Li2022BLIPBL,Li2023BLIP2BL} compresses raw video or audio into compact representations, while pre-trained LLMs \cite{Chung2022ScalingIL,vicuna2023,Touvron2023Llama2O} supply broad world knowledge for downstream tasks. Since LLMs cannot directly process encoder outputs, a learnable interface aligns the modalities, often via a Q-Former that integrates at the token \cite{Lin2023VideoLLaVALU} or feature \cite{Alayrac2022FlamingoAV} level. A more detailed description of a typical VLM architecture is given in Appendix \ref{sec:mod}.

While many models score high on popular benchmarks (not least because the benchmark data may appear in their training sets), they typically struggle with novel scenarios. As we developed a completely new benchmark and set the baseline, for evaluation we focused on top models from the HuggingFace VLM leaderboard, including Pangea \cite{yue2024pangeafullyopenmultilingual}, Gemini \cite{geminiteam2024geminifamilyhighlycapable}, PALO \cite{PALO}, Video-Llama2 \cite{damonlpsg2024videollama2}, Phi-3.5-vision \cite{abdin2024phi3technicalreporthighly}, Qwen2-VL \cite{Qwen2VL}, Llava-NEXT-Video \cite{zhang2024videoinstructiontuningsynthetic}, InternVL2 \cite{chen2024internvl}, MiniCPM-V \cite{yao2024minicpm} and DeepSeek-VL \cite{lu2024deepseekvlrealworldvisionlanguageunderstanding}. We filtered out the models that do not support the languages featured in Perfect Times.

%% file: method.tex
\section{Method}
\label{sec:method}

We anchor the main clause action, the queried action in the correct answer (\textit{mca}), and position the dependent clause action that sets the temporal and causal context relative to it (\textit{dca}). This sequential shift reflects their temporal progression and fully covers Allen's interval algebra \cite{allen1984towards} that encodes all possible temporal relationships between two actions. Linguistically, these relations map to verb tenses and aspects and subordinate temporal conjunctions: for instance, \textit{after} (Italian \textit{dopo (che)}, Russian \selectlanguage{russian}
после того, как\selectlanguage{english}, Japanese \begin{CJK}{UTF8}{min}\textit{後}\end{CJK}) indicates that the \textit{dca} precedes the \textit{mca}; \textit{while} (Italian \textit{mentre}, Russian \selectlanguage{russian}\textit{пока}\selectlanguage{english}, Japanese \begin{CJK}{UTF8}{min}\textit{ながら}\end{CJK}) denotes simultaneity; and \textit{before} (Italian \textit{prima (che)}, Russian \selectlanguage{russian}\textit{перед тем, как}\selectlanguage{english}, Japanese \begin{CJK}{UTF8}{min}\textit{前に}\end{CJK}) signals that the \textit{dca} follows the \textit{mca}.

For perfectivity, which captures an action's continuous or complete (and causally linked) nature, we focus on the start and end points of both the main (\textit{st\_mca}, \textit{et\_mca}) and dependent (\textit{st\_dca}, \textit{et\_dca}) clauses. We also label each action as telic or atelic based on its inherent properties. Accordingly, we group Allen's temporal relations into three categories (precedence, succession, and simultaneity for the main clause action) and define the plausible aspect correlations within each group. For example, an incomplete action without a clear endpoint is unlikely to occur \textit{before} another action; this is a hardly plausible combination in sequential events even if grammatically acceptable in languages like Russian.

Next, inspired by STAR \cite{Wu2021STARAB} and its sequence functional programs that exploit before/after relations, we developed templates in four languages to cover all temporal boundaries in conjunction with the semantics of action completion in both the main clause and dependent clause actions \footnote{Our goal is not to exhaustively cover all surface-level expressions of temporal relations (e.g., alternative conjunctions, adverbial phrases, or nominal constructions). Instead, we focus on evaluating the model’s overall comprehension of temporal semantics with respect to perfectivity, regardless of paraphrasing.}. The templates were developed with the help of native speakers who have linguistic or philological background and specialize in translations.

For instance, when \textit{mca} is a completed action, we use as its base \textit{What \textbf{did} the person in the video do…} (Italian Cosa ha fatto la persona nel video…/Cosa aveva fatto la persona nel video…/Cosa fece la persona nel video…; Russian \selectlanguage{russian}Что сделал человек на видео…\selectlanguage{english}; Japanese \begin{CJK}{UTF8}{min}ビデオに写っている人は…何をしましたか\end{CJK}). The ongoing action as \textit{mca} corresponds to \textit{What \textbf{was} the person in the video \textbf{doing}…?} (Italian Cosa stava facendo la persona nel video… or Cosa faceva la persona nel video…; Russian \selectlanguage{russian}Что делал человек на видео…\selectlanguage{english}; Japanese \begin{CJK}{UTF8}{min}ビデオに写っている人は…何をしていましたか\end{CJK}).

Below we outline temporal relations and templates by groups, while the complete list of templates in all languages is given in Appendix \ref{sec:templ}. The examples of all the temporal questions in Perfect Times are given in Table \ref{tab:temporal_aspectual_examples}\footnote{In the table, we provide examples in English for the general reference; some other examples in all the languages are in Appendix \ref{sec:examples}.}.

\subsection{Precedence}
\label{sec:prec}

\begin{figure}[htbp]
\centering
\includegraphics[width=\columnwidth]{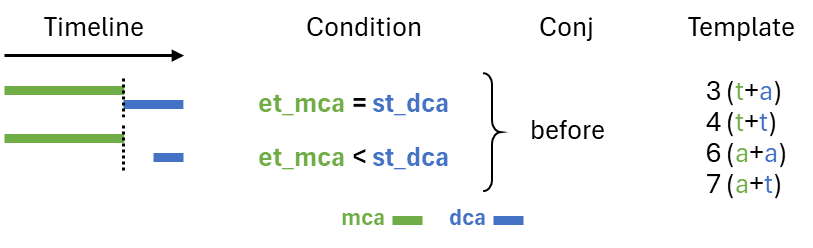}
\caption{Precedence of mca in templates 3, 4, 6, and 7 with all combinations of mca and dca telicity markers.}
\label{fig:before}
\end{figure}
As shown in Figure \ref{fig:before}, when mca happens before dca (et\_mca $\leq$ st\_dca), all combinations of action completeness are possible. In the languages with tense agreement, English and Italian, the tense of mca is backshifted to a past form relative to dca. Russian and Japanese, in contrast, convey only the aspect of both mca and dca in place of shifting tenses. 

\subsection{Succession}
\label{sec:succ}

\begin{figure}[htbp]
\centering
\includegraphics[width=\columnwidth]{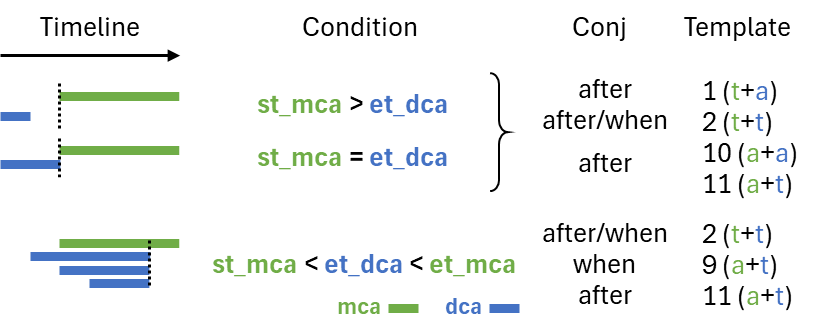}
\caption{Succession of mca in templates 1, 2, 9, 10, 11. The strict sequence of actions allows for all combinations of telicity in mca and dca. Attention is focused on st\_mca after or simultaneously with et\_dca. If mca continues after dca, the sequence of events is focused on et\_dca, which is perfective.}
\label{fig:after}
\end{figure}

When mca follows dca, the trigger is the dca's endpoint: at least some part of mca must take place after it. Symmetrically to precedence, English and Italian have agreement of tenses. 

\begin{figure}[htbp]
\centering
\includegraphics[width=\columnwidth]{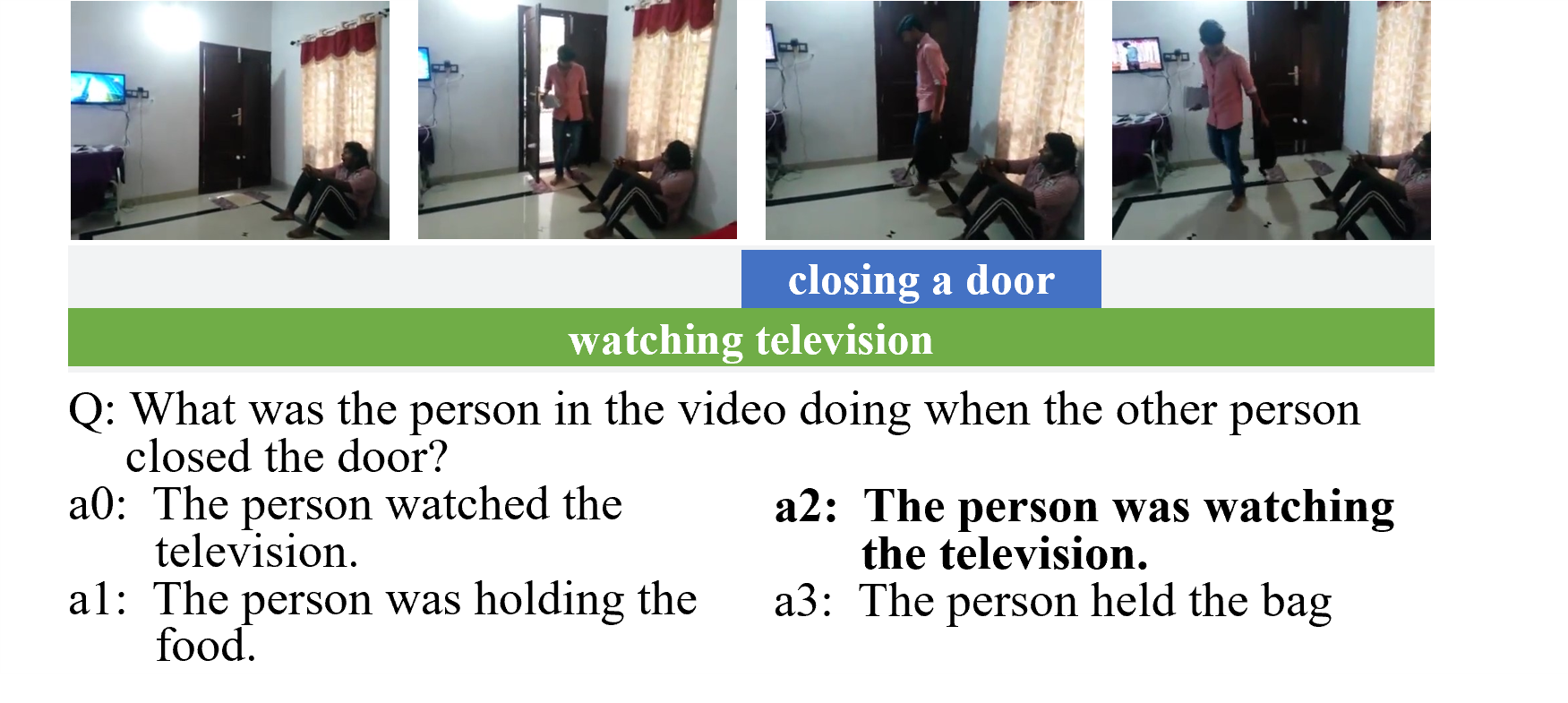}
\caption{Example in Perfect Times made by template 9b: two actions with different agents, durative mca and telic dca.}
\label{fig:temp_9}
\end{figure}

In case of partial following of the telic dca by the atelic mca, the general temporal conjunction \textit{when} (Italian quando; Russian \selectlanguage{russian}когда\selectlanguage{english}; Japanese \begin{CJK}{UTF8}{min}時\end{CJK}) is used the same way as the explicit sequential \textit{after}, even though both actions may not strictly follow one another. This relationship is coded by Template 9 (Figure \ref{fig:temp_9}).

\subsection{Simultaneity}
\label{sec:sim}

\begin{figure}[htbp]
\centering
\includegraphics[width=\columnwidth]{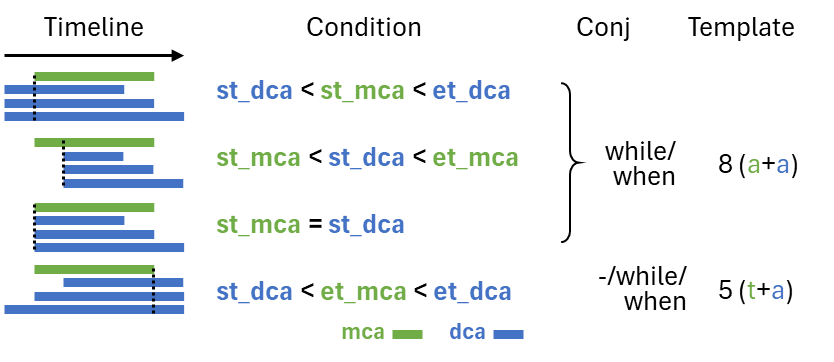}
\caption{Overlapping mca and dca in templates 5 and 8.}
\label{fig:during}
\end{figure}

The prototypical conjunctions for expressing overlapping actions, or simultaneity, are \textit{while} (Italian mentre, Russian \foreignlanguage{russian}{пока}, Japanese \begin{CJK}{UTF8}{min}ながら/間に\end{CJK}) and the more general \textit{when}. The continuous nature of dca (against which the mca of any perfectivity is questioned) underlies these relations.

\begin{figure}[htbp]
\centering
\includegraphics[width=\columnwidth]{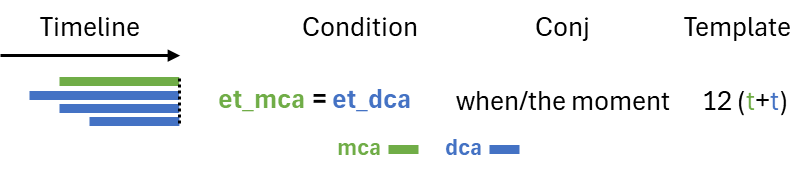}
\caption{Template 12: simultaneity when both actions mca and dca end at the same moment in.}
\label{fig:moment}
\end{figure}

When two telic actions finish at the same time, the ambiguous \textit{when} emphasizes the simultaneity of the completion – as a synonym to \textit{the moment \textbf{when}...} (Figure \ref{fig:moment}).

Each QA pair comes with three distractors that consider both the semantics and the grammatical form of the action. 

\subsection{Distractors}

\begin{figure}[htbp]
\centering
\includegraphics[width=\columnwidth]{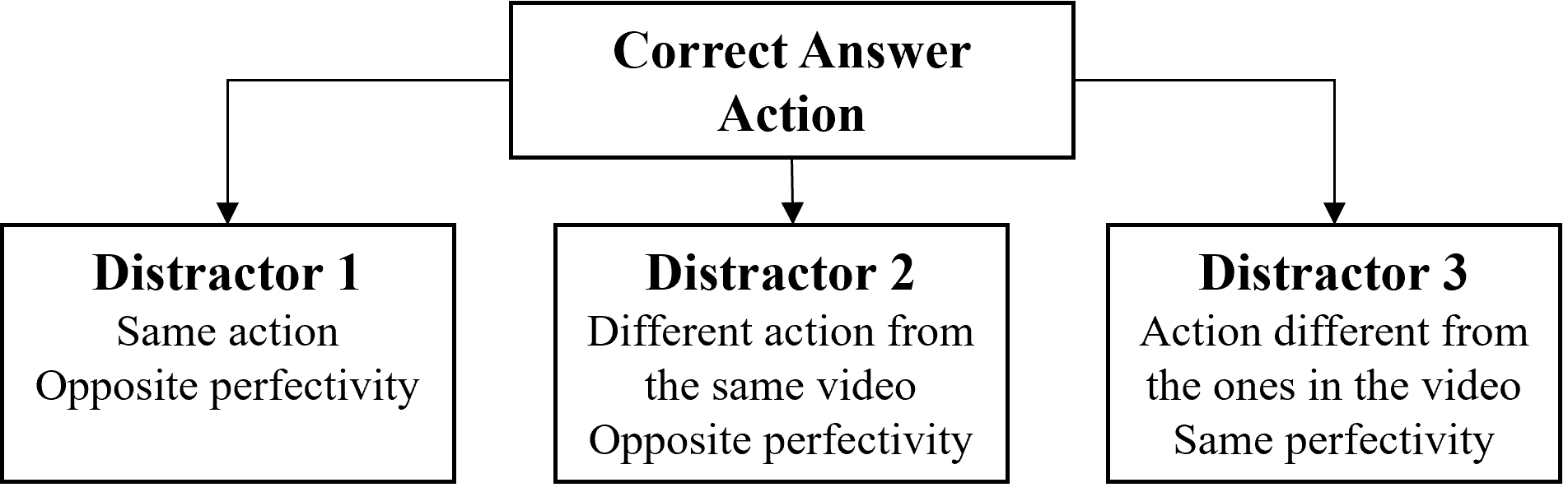}
\caption{Distractors in relation to the Correct Answer.}
\label{fig:distractors}
\end{figure}

Each distractor type is designed to probe different aspects of comprehension: linguistic nuance, context relevance, and discriminative ability within a shared scene. The diagram in Figure \ref{fig:distractors} illustrates the configuration of the distractors.

\textbf{Distractor Type 1} is a variation of the correct answer rendered in a different form to convey the opposite completeness of the action (e.g., switching from past simple tense to a continuous/ing form in English or taking the opposite aspect word in Russian). It is intended to be very close to the correct answer with only a subtle difference in linguistic details. This tests whether a model (or human) can notice and correctly interpret the aspect mismatch between the question and the answer. 

\textbf{Distractor Type 2} comes from an alternative action that occurs in the same video, but uses the switched perfectivity/duration type the same way as in Type 1. Being contextually related, it forces the evaluator to differentiate between two plausible actions within the same scene. It tests the boundary between action comprehension and grammatical understanding.

\textbf{Distractor Type 3} is taken from a completely different context. It comes from the list of actions that do not appear in the current video. Its purpose is to introduce an option that should be clearly out of context for the video at hand. A model with good action recognition should be able to dismiss such an option as completely implausible.

Among the four answer choices, both completed and durative actions are represented equally. Each answer option (one correct and three distractors) must be assigned to a random position for each question. To avoid option selection bias \cite{loginova2024addressingblindguessingcalibration}, all distractor types and the correct answer should be distributed approximately equally across the answer options.

%% file: results.tex
\section{Experiments and Results}
\label{sec:exp}

\subsection{Dataset}
\label{sec:dataset-main}

Four annotators\footnote{The annotators were recruited as volunteers and were all proficient in at least two target languages. Two of them have a background in linguistics, and their academic levels include two undergraduates and two postgraduates. Although none were directly involved in the project, each received detailed briefing and debriefing instructions from the professional linguist on the team. Their language proficiencies are as follows: Annotator 1 — English (fluent), Italian (intermediate), Russian (native), Japanese (fluent); Annotator 2 — English (native), Russian (native); Annotator 3 — English (advanced), Italian (native), Russian (native); Annotator 4 — Italian (native), Russian (native).} labeled 400 clips of the Charades \cite{Sigurdsson2016HollywoodIH} collection of approximately 30-second videos. All activities in these videos were categorized into 157 verb action classes derived from Action Genome \cite{Rai2021HomeAG}, along with their respective time boundaries\footnote{The Charades dataset was initially annotated with activities and their time intervals for different research purposes. This previous annotation was inadequate for precisely determining the sequence of events or their concurrent occurrence due to its broad definition of action boundaries and a general interpretation of actions without considering their causal and temporal relationships. Additionally, the initial markup failed to note whether different actions were performed by the same character or different ones, a crucial detail for constructing meaningful questions.}. The statistics of videos and classes are given in Appendix \ref{sec:video}.

The action classes, which are essentially verb phrases, were also annotated with telicity labels irrespective of their context. Inter-annotator agreement scores 0.67 by Fleiss' kappa, which is considered substantial \cite{Landis77}, although it indicates that there is some inherent ambiguity or subjectivity in assessing the verb classes for telicity in isolation.

The algorithm begins by mixing actions within the same video. It goes line by line in the shuffled annotation file and assigns mca to the first action and dca to the second one for each pair of neighboring actions. Since the mca and dca annotations have telicity labels and time boundaries used as conditions, the algorithm applies all corresponding templates and generates the questions and correct answer options. Then the dataset is populated with distractors. The correct answers and distractor types are balanced across the answer options. We end up with the dataset of 3,739 QA pairs. 

Subsequently, speakers of each language were asked to take this MCQA test with instructions similar to the prompts for VLMs: \textit{Choose the correct answer (a0, a1, a2, or a3) and respond only with the option key (e.g., a0)}\footnote{We deliberately did not provide additional instructions to annotators, as we aimed to collect the data based on linguistic intuition.}. As a result, we obtained an inter-annotator agreement of 0.8 according to Fleiss' kappa (substantial agreement). This is how the gold standard of 93.36\% accuracy was developed. This high percentage is due to the fact that distractors are designed in such a way that the correct answer is not difficult to intuitively predict. Notably, most annotators made the most mistakes in Distractor Type 2. Additional statistics on the annotators' responses are presented in Appendix \ref{sec:annot}.

\subsection{Models}
\label{sec:models}

We tested both the open-source and closed-source multilingual VLMs: Qwen2-VL \cite{Qwen2VL}, MiniCPM-V \cite{yao2024minicpm}, InternVL2 \cite{chen2024internvl}, LLaVA-NeXT-Video \cite{zhang2024videoinstructiontuningsynthetic}, GPT-4o \cite{Achiam2023GPT4TR}, and Gemini-2.0-Flash-Lite \cite{geminiteam2024geminifamilyhighlycapable}. Additional details on open-source models are given in Table \ref{tab:model_configs} of Appendix \ref{sec:mod}. 

We passed the video either directly (Qwen2-VL, Gemini-2.0-Flash-Lite and LLaVA-NeXT-Video) or first extracted frames every 3 seconds and passed the list of frames to those models that can only process a sequence of images (GPT-4o, InternVL2 and  MiniCPM-V). The exact prompts for each model are provided in Appendix \ref{sec:prompts}.

\subsection{Results}
\label{sec:results}

\subsubsection{Evaluation}
\label{sec:eval}
We measure \textbf{Accuracy}, \textbf{F1 Macro}\footnote{We also checked F1 Micro, but because the dataset is rather balanced, this metric with TP/FP/FN does not differ much from Accuracy.} and \textbf{Distractor Rate by Type}, which is the percentage of choices of each distractor type in case of error, calculated as follows: 

Let us define \textbf{$D_i$} — the type of distractor chosen by the model for question $i$, if it was predicted incorrectly; \textbf{$T_k$} — a specific type of distractor (e.g., Type 1, Type 2, Type 3); \textbf{$n_k$} — the number of times the model selected a distractor of type $T_k$.

The proportion of errors on distractors of type $T_k$ is calculated as:

\begin{equation}
    P(T_k) = \frac{n_k}{\sum_{j} n_j} \times 100,
\end{equation}

where \textbf{$P(T_k)$} — the probability of selecting a distractor of type $T_k$ and \textbf{$\sum_{j} n_j$} — the total number of cases where the model made an error\footnote{A similar metric can be applied if we compute the error frequency relative to all predictions, not just erroneous ones: $P(T_k) = \frac{n_k}{N} \times 100$, where $N$ is the total number of questions.}.

\subsection{Quantitative and Qualitative Results}
\label{sec:quant} 

\begin{table*}[t]
\caption{VLM performance on Perfect Times across different languages (in percentage).}
\label{tab:model_performance}
\centering
\small
\setlength{\tabcolsep}{4pt}
\renewcommand{\arraystretch}{1.2}
\begin{tabular}{lccccc}
\toprule
\textbf{Model} & \textbf{Accuracy} & \textbf{F1 Macro} & \textbf{Distractor Type 1} & \textbf{Distractor Type 2} & \textbf{Distractor Type 3} \\
\midrule
\multicolumn{6}{c}{\textbf{English}} \\
\midrule
Gemini-2.0-flash-lite & 43.41 & 42.19 & 22.15 & 30.42 & 4.01 \\
GPT-4o & 43.25 & 34.39 & 21.96 & 31.14 & 3.64 \\
MiniCPM-V-2\_6 & 36.19 & 35.5 & 27.68 & 31.08 & 5.05 \\
Qwen2-VL-7B-Instruct & 35.09 & 32.68 & 21.24 & 39.2 & 4.47 \\
InternVL2-8B & 34.86 & 33.36 & 27.34 & 29.09 & 8.7 \\
LLaVA-NeXT-Video-7B & 33.38 & 32.86 & 25.39 & 30.99 & 10.22 \\
\midrule
\multicolumn{6}{c}{\textbf{Italian}} \\
\midrule
Gemini-2.0-flash-lite & 43.11 & 42.15 & 21.85 & 30.78 & 4.25 \\
Qwen2-VL-7B-Instruct & 41.59 & 39.63 & 21.95 & 32.09 & 4.35 \\
GPT-4o & 40.71 & 32.34 & 25.29 & 31.45 & 2.49 \\
MiniCPM-V-2\_6 & 37.42 & 35.73 & 29.55 & 28.96 & 4.07 \\
InternVL2-8B & 34.07 & 32.38 & 32.00 & 24.63 & 9.3 \\
LLaVA-NeXT-Video-7B & 25.92 & 17.65 & 25.40 & 29.74 & 18.88 \\
\midrule
\multicolumn{6}{c}{\textbf{Russian}} \\
\midrule
Gemini-2.0-flash-lite & 46.99 & 46.33 & 19.04 & 29.85 & 4.12 \\
GPT-4o & 45.04 & 44.97 & 19.60 & 32.15 & 3.21 \\
InternVL2-8B & 36.87 & 34.37 & 28.82 & 25.54 & 8.77 \\
MiniCPM-V-2\_6 & 36.29 & 34.61 & 28.88 & 30.20 & 4.63 \\
Qwen2-VL-7B-Instruct & 34.13 & 32.75 & 20.40 & 39.5 & 5.96 \\
LLaVA-NeXT-Video-7B & 26.95 & 24.7 & 24.3 & 36.00 & 12.74 \\
\midrule
\multicolumn{6}{c}{\textbf{Japanese}} \\
\midrule
Gemini-2.0-flash-lite & 43.06 & 41.77 & 22.19 & 31.43 & 3.32 \\
MiniCPM-V-2\_6 & 38.73 & 36.31 & 30.25 & 26.34 & 4.68 \\
GPT-4o & 38.49 & 30.65 & 23.78 & 35.23 & 2.49 \\
Qwen2-VL-7B-Instruct & 37.52 & 36.85 & 20.17 & 37.90 & 4.41 \\
InternVL2-8B & 35.05 & 31.65 & 31.22 & 23.92 & 9.81 \\
LLaVA-NeXT-Video-7B & 26.18 & 19.24 & 25.25 & 33.06 & 15.41 \\
\bottomrule
\end{tabular}
\end{table*}

A total of 5.67\% (208)\footnote{See detailed statistics in Appendix~\ref{sec:corr}.} of questions were answered incorrectly by all models. These correspond to 162 videos, with only one video where every question was answered incorrectly by all models\footnote{This could suggest problems in question formulation, but a quick manual check of 10\% of these questions confirmed they are unambiguously answerable by humans.}. Thus, the dataset contains no videos completely unintelligible to all models; instead, failures are localized to specific actions or the correct order of actions within certain videos. The accuracy of all models in all languages remains significantly below the human gold standard. Coupled with consistent error trends across models and languages, this highlights that even the strongest current VLMs lack robust mechanisms for precise temporal fusion.

Table \ref{tab:model_performance} shows the results of our tests on Perfect Times dataset with the breakdown by the distractor types. All models tend to ignore linguistic aspect, as most errors are related to Distractor Types 1 and 2, where the predicted verb form does not match the verb form in the question. The majority of errors, except for InternVL2, are due to Type 2 distractors. This suggests that models do not identify which temporal moment is crucial for the answer, even if they can distinguish actions relevant to the video. Therefore, VLMs should leverage multimodal fusion or specialized architectures to better capture temporality.

InternVL2, in contrast, demonstrates good understanding of temporal context but struggles with subtle linguistic differences in aspect; its majority of errors are Type 1 distractors in every language except English (possibly due to greater data contamination). Its worst performance in Italian (the language with the most diverse verb forms for encoding temporal relations) points to the model's lack of understanding of grammatical distinctions even though it is better at localizing events with respect to their order in the video. 

Across all languages, LLaVA-NeXT-Video performs the worst. Despite its multilingual LLM backbone (based on Qwen \cite{qwen}), its performance is nearly random for languages other than English. Moreover, answer analysis revealed a severe selection bias\footnote{All other models revealed less prominent selection bias in the first option, a0.} towards the first and second options in English and the first option in other languages. This indicates a systematic flaw in its logic and temporal reasoning, even though it is able to handle basic contextual cues (Distractor Type 3 is a minority). Its temporal capabilities could be improved through enhanced multilingual action recognition of spatiotemporal features.

Gemini-2.0-flash-lite consistently performed better across all languages, showing similar accuracy in English, Italian, and Japanese. Other models display greater variance in accuracy. A thorough causal analysis for this model would require access to training data and architecture, which is not possible; however, factors such as training data saturation and tokenizer limitations (particularly in morphologically diverse languages such as Italian and Russian) likely contribute to the errors. 

Gemini-2.0-flash-lite, GPT-4o, and InternVL2-8B generally perform better on Russian data, where perfectivity is lexically encoded, and worse on Japanese, which demands stronger visual disambiguation. This reinforces our claim in Section~\ref{sec:intro} that language-specific encoding of time and aspect strongly affects temporal reasoning. It also echoes findings in the cognitive literature discussed in Section~\ref{sec:cog}, where perfectivity encoded via telicity is more readily interpreted by native speakers.

Another parallel between human cognitive processes of causality and model behavior emerges from qualitative analysis: all tested VLMs, when mistaken, prefer perfect (or telic) answers over durative (or atelic) ones\footnote{Also, considering correct answers only, GPT-4o is the only model that predicts fewer correct telic classes. While this cannot be confirmed for closed-source models, we tentatively attribute it to the volume and contamination in their training data. Table~\ref{tab:telicity_accuracy} in Appendix~\ref{sec:corr} presents the telicity breakdown.}. This may reflect the characteristics of source data, where strong causal connections between sequential actions are prevalent.

Regarding specific temporal conjunctions, all tested models struggle to disambiguate \textit{when} (and equivalents in other languages). The highest number of correct answers is found for simultaneity, marked with \textit{while} (Template 8a1), whereas replacing \textit{while} with \textit{when} for the same temporal relation and identical verb forms (Template 8a2) results in significantly more errors\footnote{Appendix~\ref{sec:conj} provides examples and quantitative data.}. Similarly, the unambiguous conjunction \textit{after} in 2a1 and 2b1 yields fewer errors than 2a2 and 2b2, which use \textit{when} to describe the same action. This behavior highlights a weakness in aligning temporal boundaries and the semantics of action relations with verb aspect. Following cognitive findings that underline human reliance on integrated multimodal cues, it indicates that the separate language stream in multimodal models, even when combined with vision, is insufficient for temporal reasoning in videos.

The best-answered templates concern \textit{after} questions, while \textit{before} questions are the most challenging. When processing visual information, models more easily identify subsequent actions than preceding ones. This is further evidenced by the prevalence of Type 2 distractor errors, where the action is present in the video, but not in the correct sequence or form. Therefore, model errors stem not from a lack of understanding of the entire video or individual actions, but from a misinterpretation of the temporal order, such as answering about a subsequent action when asked about a previous one.

This pattern may also result from imbalanced instruction-tuned training data for temporal benchmarks. According to psychological research, it is more natural for people to describe events in chronological order via causal relationships, rather than discuss the reverse order\footnote{This is also confirmed by the NeXT-QA and STAR datasets, widely used for causal and temporal evaluation: the overwhelming majority of temporal questions focus on succession via \textit{after}, rather than precedence via \textit{before}.}.

Another notable finding is that the largest number of questions answered correctly by all five models except LLaVA-NeXT-Video-7B\footnote{LLaVA-NeXT-Video-7B is excluded because its results for non-English languages are statistically non-informative and nearly random.} is in Italian, and the smallest in Japanese (see Tables \ref{tab:all_correct_lang}-\ref{tab:all_incorrect_lang} in Appendix~\ref{sec:corr}). This suggests that explicit surface syntactic information can help in choosing correct answers and establishing proper temporal relations, provided it is systematically integrated into large-scale statistical processing, even when the dataset contains noise.

%% file: conclusion.tex
\section{Conclusion}
\label{sec:concl}

We have demonstrated that a full understanding of event dynamics requires an integrated approach where both visual cues and grammatical structures inform the interpretation of action completion. Our cross-linguistic evaluation shows that the SoTA VLMs rely on limited superficial cues, particularly when disambiguating ambiguous temporal conjunctions and distinguishing subtle aspectual markers. Models favor telic responses, mirroring human tendencies to find cause and effect dependencies in events and exposing a reliance on lexical over grammatical signals. These observations reinforce the need for enhanced multimodal fusion and specialized temporal representations.

A key advantage of our \textbf{Perfect Times} dataset is its semi-synthetic, template-based design. By leveraging a unique methodology that uses universal temporal and aspectual templates, our dataset can be easily augmented and adapted to any language, provided videos are annotated with action labels and timestamps. This flexibility makes our benchmark not only a novel tool for evaluating temporal reasoning in VLMs but also a scalable resource to drive further research. Our work thus sets a new standard for probing the intricate interplay of time, causality, and aspect in multimodal contexts.

%% file: limitations.tex
\section{Limitations}
\label{sec:lim}

Our study is the first to link tense, aspect, and visual modality to test deep temporal reasoning in models. However, it is not exhaustive regarding the full range of synonymous temporal constructions. Future work could augment the dataset with paraphrases generated by LLMs to increase coverage. 

Additionally, our semi-synthetic, template-based approach may not fully reflect the variability found in natural language. Although templates ensure controlled testing of specific temporal relations, they might not fully reflect spontaneous speech. Expanding the dataset to include more naturalistic examples could provide further insights.

Another limitation is the number of verb classes: 157 classes for 400 videos. While a broader range of classes would capture more nuanced actions, the current quantity is balanced by high-quality annotations and language-specific templates, which were meticulously crafted by experts, a resource-intensive process.

Finally, with more annotations per language, the gold standard accuracy might decrease when averaged across annotators, although it is unlikely to reach the performance levels of current state-of-the-art models. Future research should explore scaling up the dataset, both in terms of linguistic variety and video domains, to fully assess and enhance multimodal temporal reasoning.

%% file: latex/appendix.tex
\label{sec:appendix}

\section{Video and Action Classes Statistics}
\label{sec:video}

Tables \ref{tab:video_statistics} and \ref{tab:action_metrics_comparison} provide details on videos and action classes in Perfect Times, respectively. 

\begin{table}
  \centering
  \caption{General Statistics of Perfect Times}
  \begin{tabular}{lr}
    \hline
    \textbf{Parameter} & \textbf{Value} \\
    \hline
    Total videos & 400 \\
    Total videos duration (sec) & 11386.65 \\
    Total videos duration (min) & 189.78 \\
    Average video duration (sec) & 29.20 \\
    Minimum video duration (sec) & 7.21 \\
    Maximum video duration (sec) & 54.12 \\
    \hline
  \end{tabular}
  \label{tab:video_statistics}
\end{table}

\begin{table}
\caption{Action Annotations}
\begin{center}
\begin{tabular}{ l c }
\hline
\textbf{Parameter} & \textbf{Annotation} \\
\hline
Total actions & 3115 \\
Minimum actions per video & 2 \\
Maximum actions per video & 30 \\
Average actions per video & 7.99 \\
Average time per action (sec) & 8.35 \\
Actions with $\leq$ 1s duration & 1149 (36.89\%) \\
Actions with $\leq$ 2s duration & 1437 (46.13\%) \\
Actions with $\leq$ 5s duration & 1850 (59.39\%) \\
Actions with $\leq$ 10s duration & 2259 (72.52\%) \\
\hline
\end{tabular}
\end{center}
\label{tab:action_metrics_comparison}
\end{table}

\section{Dataset Statistics}
\label{sec:dat_stat}

The breakdown by answer options with the correct answers and distractor types of the total of 3739 qa pairs is given in Table\ref{tab:qa_stats}

\begin{table}[h!]
\centering
\small
\setlength{\tabcolsep}{6pt}
\renewcommand{\arraystretch}{1.2}
\caption{QA pair statistics and distractor types.}
\label{tab:qa_stats}
\begin{tabular}{lc}
\toprule
\textbf{Parameter} & \textbf{Value} \\
\midrule
Total QA pairs  & 3739 \\
Correct a0 & 953 \\
Correct a1 & 966 \\
Correct a2 & 960 \\
Correct a3 & 860 \\
a0\_distractor\_type & 953 \\
a1\_distractor\_type & 916 \\
a2\_distractor\_type & 927 \\
a3\_distractor\_type & 943 \\
\bottomrule
\end{tabular}
\end{table}

\section{Templates}
\label{sec:templ}

This section presents templates in all languages, taking into account the parsed verb phrases from video annotations. In most cases, the classes include one verb, verb\_1. In the tables below for several verbs in one class in coordinating constructions, conj, the coordinating conjunction and all elements with \_2 represent the second conjunct, as in \textit{working \textbf{or playing on the laptop}}. The second conjunt in the verb phrase is optional.

\section{Dataset Examples}
\label{sec:examples}

\begin{figure*}[htbp]
\centering
\includegraphics[width=\textwidth]{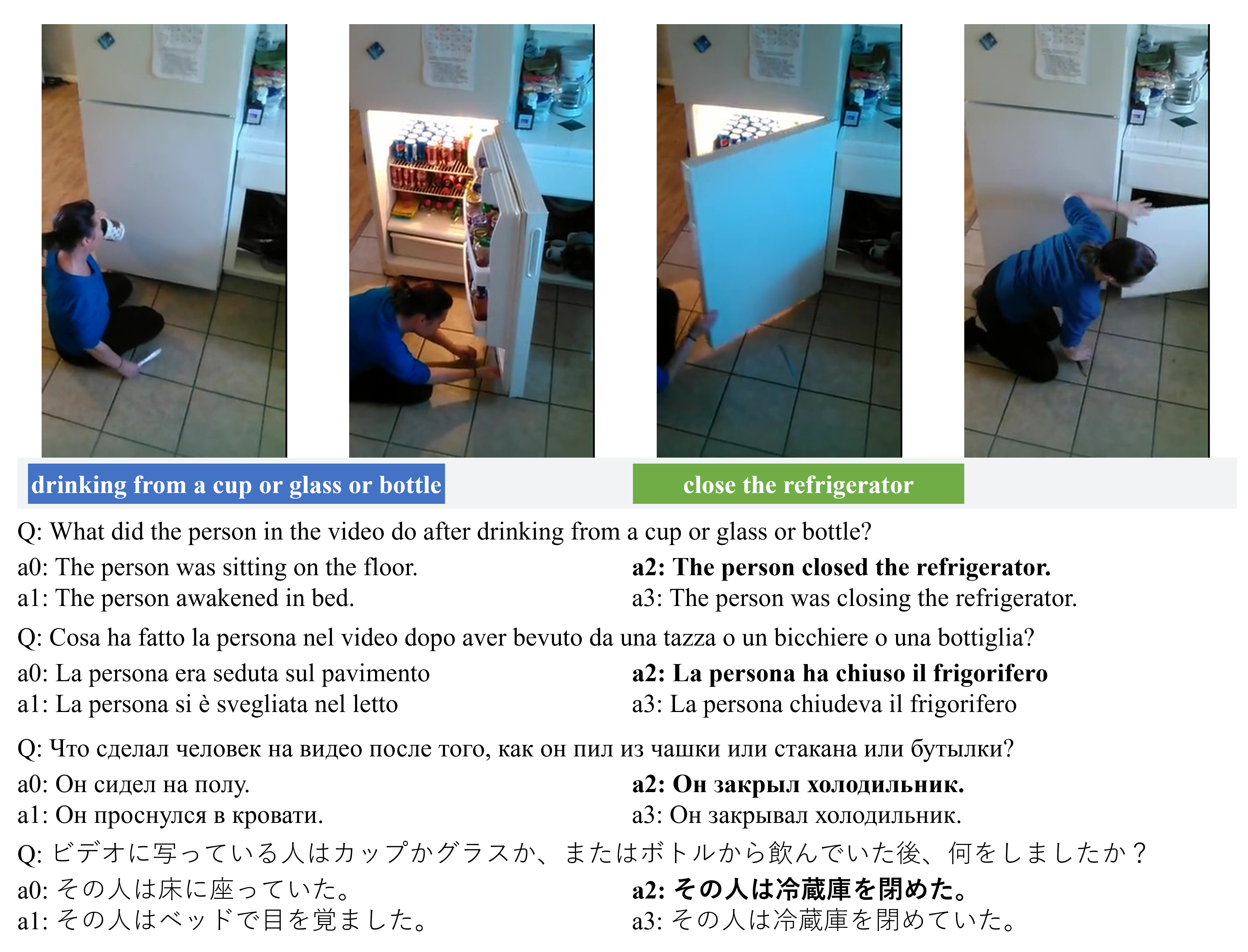}
\caption{Example from Perfect Times generated by Template 1a for all the languages.}
\label{fig:1a_ALL}
\end{figure*}

\begin{figure*}[htbp]
\centering
\includegraphics[width=\textwidth]{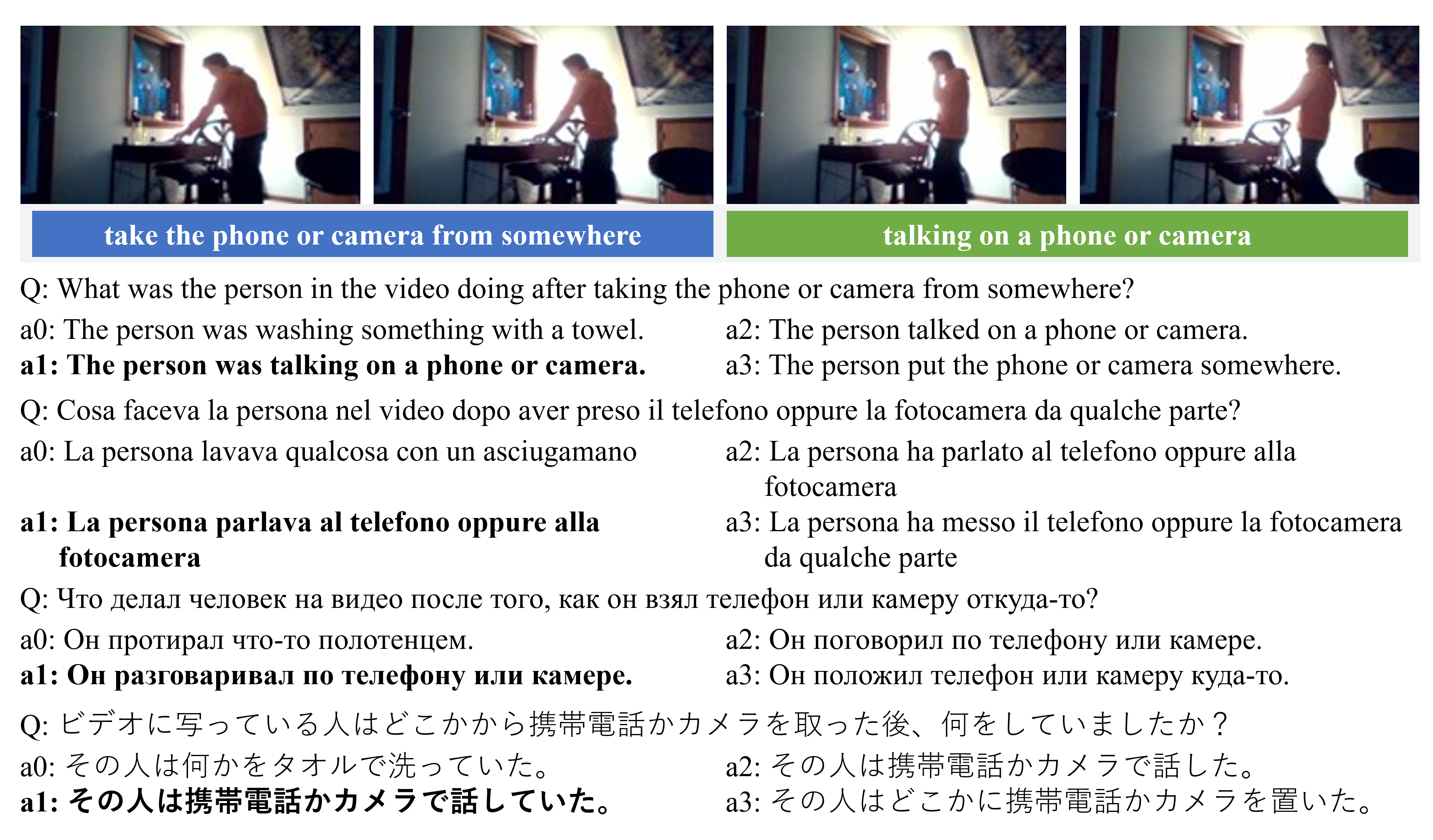}
\caption{Example from Perfect Times generated by Template 11a for all the languages.}
\label{fig:11a_ALL}
\end{figure*}

\begin{figure*}[htbp]
\centering
\includegraphics[width=\textwidth]{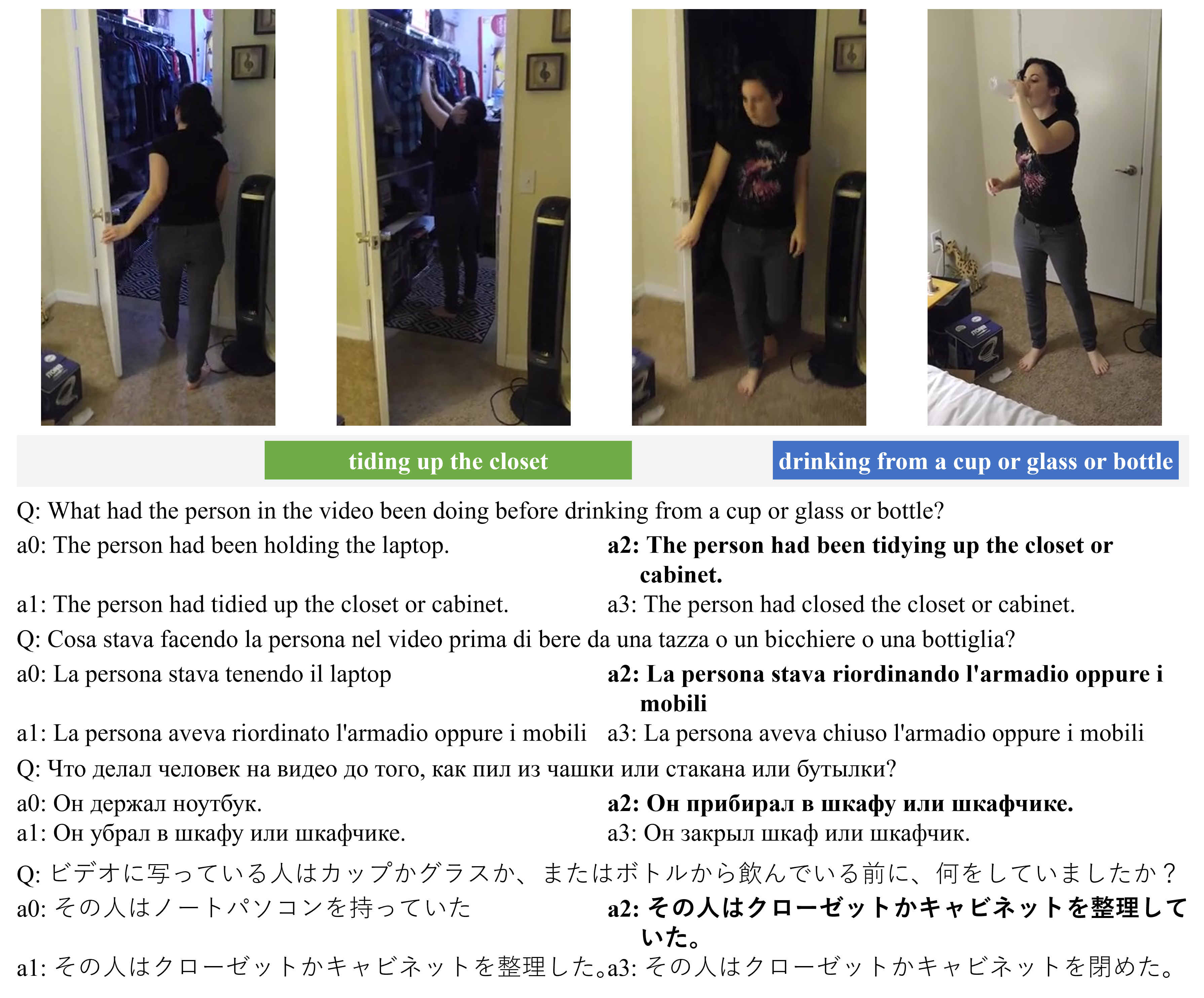}
\caption{Example from Perfect Times generated by Template 6a for all the languages.}
\label{fig:6a_ALL}
\end{figure*}

\begin{figure*}[htbp]
\centering
\includegraphics[width=\textwidth]{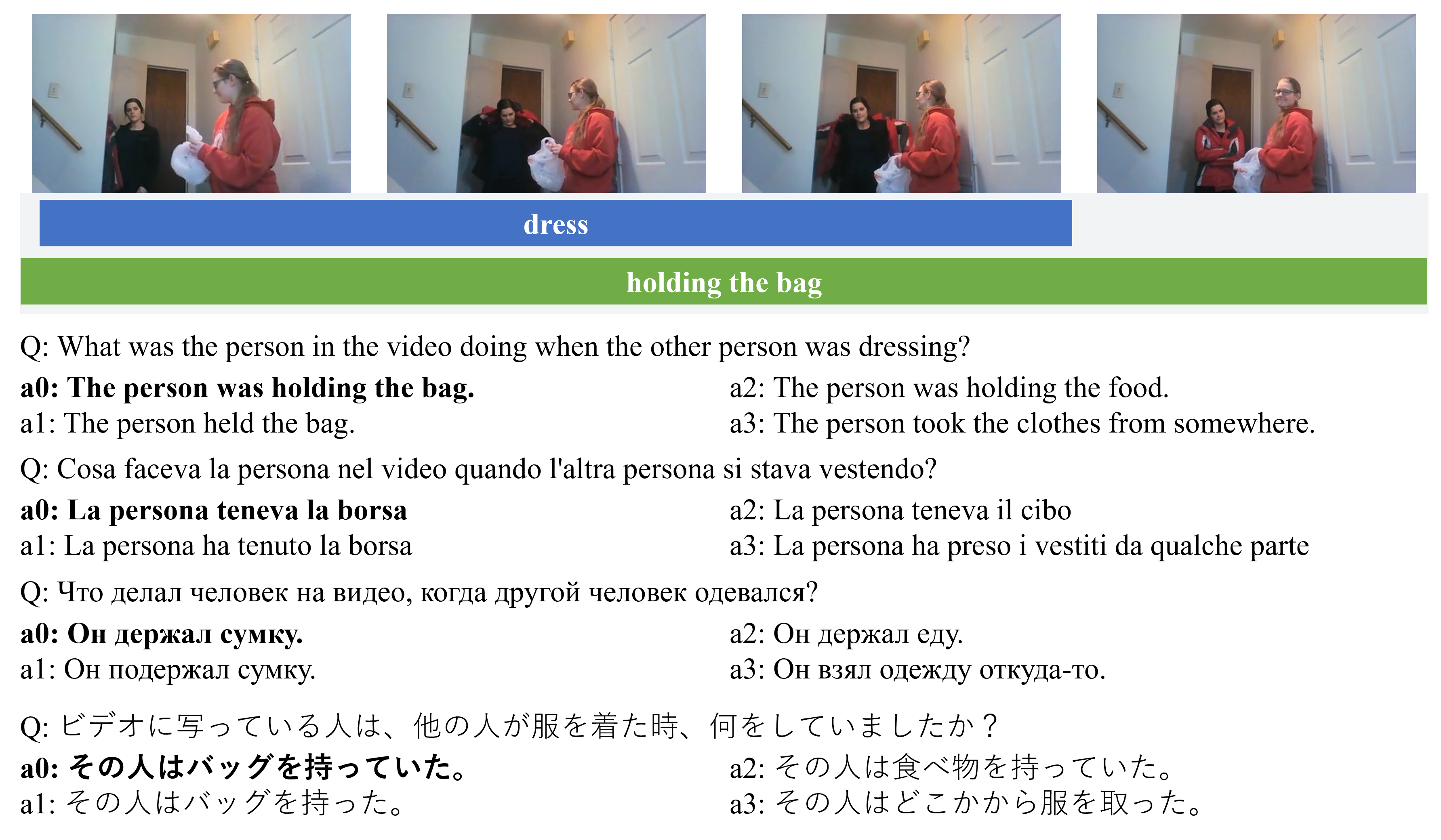}
\caption{Example from Perfect Times generated by Template 8b2 for all the languages.}
\label{fig:8b2_ALL}
\end{figure*}

\begin{figure*}[htbp]
\centering
\includegraphics[width=\textwidth]{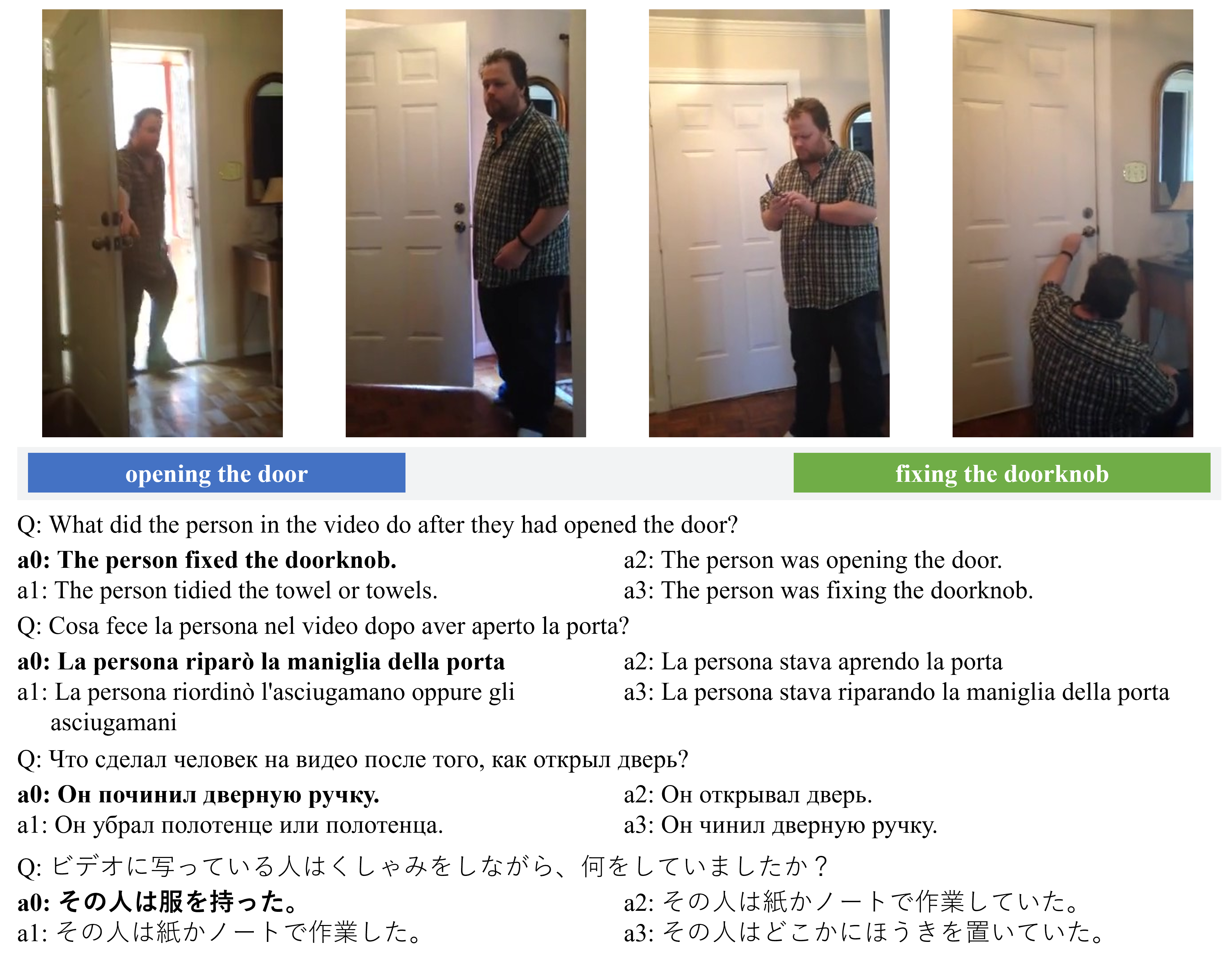}
\caption{Example from Perfect Times generated by Template 2a1 for all the languages.}
\label{fig:2a1_ALL}
\end{figure*}

\begin{figure*}[htbp]
\centering
\includegraphics[width=\textwidth]{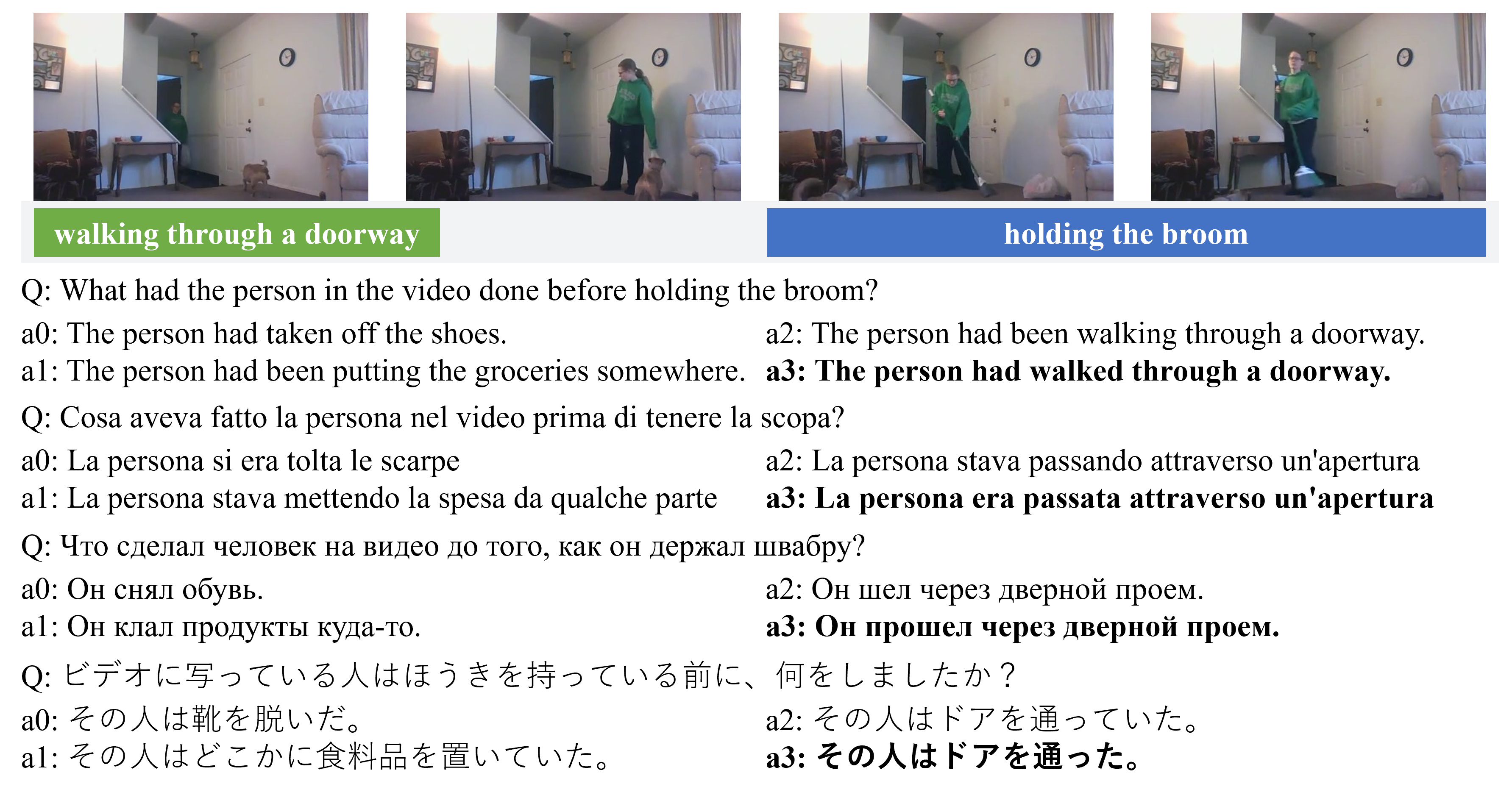}
\caption{Example from Perfect Times generated by Template 3a for all the languages.}
\label{fig:3a_ALL}
\end{figure*}

The aggregated examples in all languages are presented in Figures \ref{fig:1a_ALL} - \ref{fig:3a_ALL}.

\section{Annotator's Statistics}
\label{sec:annot}

The statistics on each annotator's performance is given in Table \ref{tab:annotator_stats}, where each language represents the annotator of this language. 

\section{Models}
\label{sec:mod}

The core idea of VLMs is to bridge the gap between visual features extracted from video and linguistic features from text for such tasks as video captioning, video question answering, video summarization, and more.

Most VLMs typically share a high-level architecture comprising three main components:

\begin{enumerate}
    \item Visual Encoder: Responsible for extracting meaningful features from the video stream.
    \item Language Encoder (and often a Language Decoder): An LLM that handles textual input and generates textual output.
    \item
    Multimodal Fusion Projector: This component connects the visual and language parts, allowing them to interact and produce a unified representation. After extracting features from individual frames, it maps these visual features into the same embedding space as the language model's tokens.
\end{enumerate}

A fundamental aspect of VLM processing is that video is generally treated as a sequence of individual images (frames). To prevent the loss of temporal information when processing each frame independently, VLMs often incorporate mechanisms to capture motion and temporal relationships, such as with concatenation of features or through a dedicated temporal transformer.

Frames are sampled either uniformly or by identifying keyframes. For models like GPT-4o, InternVL2, and MiniCPM-V, we use uniform sampling by extracting frames at regular intervals (e.g., every 3 seconds or up to an upper limit). Similarly, Gemini-2.0-Flash-Lite, while inherently multimodal, gets individual frames when it is passed the entire video and interacted with via API. In contrast, LLaVA-NeXT-Video handles video processing internally with its specialized LlavaNextVideoProcessor and Qwen2-VL uses AutoProcessor and Qwen2VLForConditionalGeneration. 

\begin{figure}[htbp]
\centering
\includegraphics[width=0.45\textwidth]{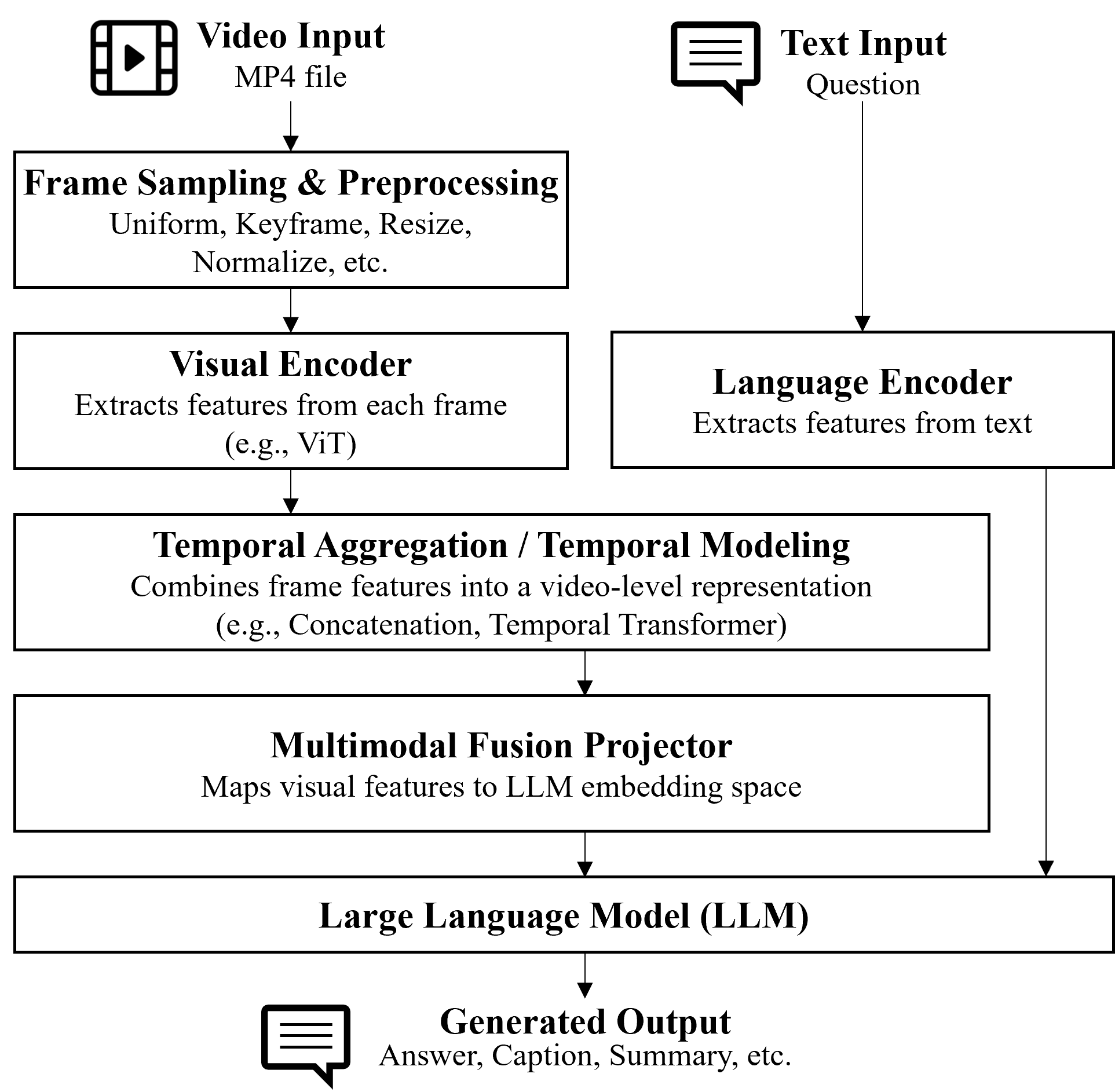}
\caption{Overview of a Typical VLM Architecture from raw video and text inputs through encoding, multimodal fusion, and final text generation.}
\label{fig:vlm}
\end{figure}

The Figure \ref{fig:vlm} illustrates the processing pipeline from multimodal input to text generation. 

Table \ref{tab:model_configs} gives details on the open-source model's components.

\section{Prompts}
\label{sec:prompts}

Since each model accepts input differently, the general question-answer options prompts have been customized.

\paragraph{Qwen2-VL Prompt}

The Qwen2-VL question messages for all the languages are straightforward, as follows: 

"en":
        question\_str = f"Question: \{question\}"
        instruction = "Choose the correct answer (a0, a1, a2, or a3)."
        
"it":
        question\_str = f"Domanda: \{question\}"
        instruction = "Scegli la risposta corretta (a0, a1, a2 o a3)."
        
"ru":
        question\_str = f"\selectlanguage{russian}Вопрос\selectlanguage{english}: \{question\}"
        instruction = "\selectlanguage{russian}Выберите правильный ответ (a0, a1, a2 или a3).\selectlanguage{english}"
        
"jp":
        question\_str = f"\begin{CJK}{UTF8}{min}質問\end{CJK}: \{question\}"
        instruction = "\begin{CJK}{UTF8}{min}正しい答えを選んでください (a0, a1, a2, a3)。\end{CJK}"

Together with the answer options they form text prompts:

{\footnotesize
\begin{verbatim}
text\_prompt = (
    f"{question_str}\n"
    f"a0: {options['a0']}\n"
    f"a1: {options['a1']}\n"
    f"a2: {options['a2']\n}"
    f"a3: {options['a3']\n}"
    f"{instruction}"
\end{verbatim}
}

The combined input to the models includes videos as follows: 
{\footnotesize
\begin{verbatim}
messages = [
    {"role": "user",
    "content": [
        {"type": "video",
        "video": video_uri,
        "fps": 1.0,
        "max_pixels": 360 * 420},
        {"type": "text", "text": text_prompt},
        ]
    }
]
\end{verbatim}
}

\paragraph{MiniCPM-V Prompt}

The text prompt for MiniCPM-V needs specific clarification about the option key. Otherwise it generates an unparsable answer. 

"en":
        question\_str = f"Question: \{question\}"
        
            instruction = "Choose the correct answer (a0, a1, a2, or a3) and respond only with the option key (e.g., a0). Do not include any additional text."
    
"it":
        question\_str = f"Domanda: \{question\}"
        
            instruction = "Scegli la risposta corretta (a0, a1, a2 o a3) e rispondi solo con la chiave dell'opzione (es. a0). Non includere testo aggiuntivo."
    
"ru":
        question\_str = f"\selectlanguage{russian}Вопрос\selectlanguage{english}: \{question\}"
        
            instruction = "\selectlanguage{russian}Выберите правильный ответ (a0, a1, a2 или a3) и ответь только ключом варианта (например, a0). Не добавляй дополнительный текст\selectlanguage{english}."
    
"jp":
        question\_str = f"\begin{CJK}{UTF8}{min}質問\end{CJK}: \{question\}"
        
            instruction = "\begin{CJK}{UTF8}{min}正しい答えを選んでください (a0, a1, a2, a3) そしてオプションキーのみで回答してください (例: a0)。追加のテキストを含めないでください。\end{CJK}"

The entire text prompt is as follows: 
{\footnotesize
\begin{verbatim}
text_prompt = (
        f"{question_str}\n"
        f"a0: {options['a0']\n}"
        f"a1: {options['a1']\\n}"
        f"a2: {options['a2']\\n}"
        f"a3: {options['a3']\\n}"
        f"{instruction}")
\end{verbatim}
}        

The message content is defined as a list that concatenates the list of video frames (as PIL images) with the text prompt the following way: 
{
\footnotesize
\begin{verbatim}
[{"role": "user", 
    "content": frames + [text_prompt]}]
\end{verbatim}
}

\paragraph{InternVL2 Prompt}

The prompt for InternVL2 contains a system message, frame intro, question, answer options and instruction. 

{
\footnotesize
\begin{verbatim}
    text_prompt = (
        f"{system_messages[language]}\n"
        f"{frame_intro[language]}\n\n"
        f"{question_str}\n"
        f"a0: {options['a0']}\n"
        f"a1: {options['a1']}\n"
        f"a2: {options['a2']}\n"
        f"a3: {options['a3']}\n"
        f"{instruction}"
    )
\end{verbatim}
}

Localized messages for instructions to InternVL2 are as follows: 

system\_messages = \{

    "en": "Use the provided video frames
    to answer the question.",
    
    "it": "Utilizza i frame del video 
    per rispondere alla domanda.",
    
    "ru": "\selectlanguage{russian}Используй данные кадры видео, 
    чтобы ответить на вопрос.\selectlanguage{english}",
    
    "jp": "\begin{CJK}{UTF8}{min}提供されたビデオフレームを使用して
    質問に答えてください。\end{CJK}"
    
\}

frame\_intro = \{

    "en": "These are the frames from the video.",
    
    "it": "Questi sono i frame del video.",
    
    "ru": "\selectlanguage{russian}Это кадры из видео.\selectlanguage{english}",
    
    "jp": "\begin{CJK}{UTF8}{min}以下はビデオのフレームです。\end{CJK}"

\}

"en":
        question\_str = f"Question: \{question\}"
        
        instruction = "Choose the correct answer (a0, a1, a2, or a3)."

"it":
        question\_str = f"Domanda: \{question\}"
        
        instruction = "Scegli la risposta corretta (a0, a1, a2 o a3)."

"ru":
        question\_str = f"\selectlanguage{russian}Вопрос:\selectlanguage{english} \{question\}"
        
        instruction = "\selectlanguage{russian}Выберите правильный ответ (a0, a1, a2 или a3).\selectlanguage{english}"

"jp":
        question\_str = f"\begin{CJK}{UTF8}{min}質問:\end{CJK} \{question\}"
        
        instruction = "\begin{CJK}{UTF8}{min}正しい答えを選んでください (a0, a1, a2, a3)。\end{CJK}"
    
\paragraph{LLaVA-NeXT-Video Prompt}

LLaVA-NeXT-Video accepts videos as input: 

{
\footnotesize
\begin{verbatim}
{
    "role": "user",
    "content": [
        {"type": "text", "text": text_prompt}, 
            {"type": "video", "video": f"file: {video_path}"}
    ]
}
\end{verbatim}
}

The text prompt has the localized question, answer options and instruction as follows: 

{
\footnotesize
\begin{verbatim}
text_prompt = (
        f"{question_str} {question}\n"
        f"a0: {options['a0']}\n"
        f"a1: {options['a1']}\n"
        f"a2: {options['a2']}\n"
        f"a3: {options['a3']}\n"
        f"{instruction}"
    )
\end{verbatim}
}

The question messages and instructions in four languages are as follows: 

translations = \{

        "en": ("Question:", "Please choose the correct answer (a0, a1, a2, or a3)."),
        
        "it": ("Domanda:", "Per favore, scegli la risposta corretta (a0, a1, a2 o a3)."),
        
        "ru": ("\selectlanguage{russian}Вопрос:", "Пожалуйста, выберите правильный ответ (a0, a1, a2 или a3).\selectlanguage{english}"),
        
        "jp": ("\begin{CJK}{UTF8}{min}質問:", "正しい答えを選んでください (a0, a1, a2, a3)。\end{CJK}")
        
    \}

\paragraph{GPT-4o Prompt}

Localized messages for making the GPT-4o prompt are as follows:

system\_messages = \{
    "en": "Use the provided video frames to answer the question.",
    
    "it": "Utilizza i frame del video per rispondere alla domanda.",
    
    "ru": "\selectlanguage{russian}Используй данные кадры видео, чтобы ответить на вопрос.\selectlanguage{english}",
    
    "jp": "\begin{CJK}{UTF8}{min}提供されたビデオフレームを使用して質問に答えてください。\end{CJK}"
\}

frame\_intro = \{
    "en": "These are the frames from the video.",
    
    "it": "Questi sono i frame del video.",
    
    "ru": "\selectlanguage{russian}Это кадры из видео.\selectlanguage{english}",
    
    "jp": "\begin{CJK}{UTF8}{min}以下はビデオのフレームです。\end{CJK}"
\}

To avoid generating multiple tokens, we further restrict the model in the instruction:

"en": (
        "Question",
        "Choose the correct answer (a0, a1, a2, or a3) and respond **only** with the option key (e.g., a0). Do not include any additional text."
        ),
        
"it": (
        "Domanda",
        "Scegli la risposta corretta (a0, a1, a2 o a3) e rispondi **solo** con la chiave dell'opzione (es. a0). Non includere testo aggiuntivo."
        ),
    
"ru": (
        \selectlanguage{russian}"Вопрос",
        "Выбери правильный ответ (a0, a1, a2 или a3) и ответь **только** ключом варианта (например, a0). Не добавляй дополнительный текст.\selectlanguage{english}"
        ),

"jp": (
        "\begin{CJK}{UTF8}{min}質問",
        "正しい答えを選んでください (a0, a1, a2, a3) そしてオプションキーのみで回答してください (例: a0)。追加のテキストを含めないでください。\end{CJK}"

The entire text prompt contains a question label, a question, answer options and an insruction as follows: 

{
\footnotesize
\begin{verbatim}
text_prompt = (
    f"{question_label}: {question}\n"
    f"a0: {options['a0']}\n"
    f"a1: {options['a1']}\n"
    f"a2: {options['a2']}\n"
    f"a3: {options['a3']}\n"
    f"{instruction}"
)
\end{verbatim}
}

In the input message to the model we pass the text prompt and frames as follows: 

{
\footnotesize
\begin{verbatim}
messages = [
        {"role": "system", "content": system_message},
        {"role": "user", "content": [frame_message, 
        *frames_payload, prompt]}
    ]
\end{verbatim}
}

\paragraph{Gemini-2.0-Flash-Lite Prompt}

Similarly to GPT-4o, the question previews and instructions for Gemini-2.0-Flash-Lite are localized with the restriction for output tokens as follows: 

"en":
        
        question\_str = f"Question: \{question\}"
        
        instruction = ("Choose the correct answer (a0, a1, a2, or a3) and respond only with the option key (e.g., a0). Do not include any additional text.")

"it":
        
        question\_str = f"Domanda: \{question\}"
        
        instruction = ("Scegli la risposta corretta (a0, a1, a2 o a3) e rispondi solo con la chiave dell'opzione (es. a0). Non includere testo aggiuntivo.")
        
"ru":
        
        question\_str = f\selectlanguage{russian}Вопрос\selectlanguage{english}: \{question\}"
        
        instruction = ("\selectlanguage{russian}Выбери правильный ответ (a0, a1, a2 или a3) и ответь только ключом варианта (например, a0). Не добавляй дополнительный текст.\selectlanguage{english}")
        
"jp":
        
        question\_str = f"\begin{CJK}{UTF8}{min}質問\end{CJK}: \{question\}"
        
        instruction = ("\begin{CJK}{UTF8}{min}正しい答えを選んでください (a0, a1, a2, a3) そしてオプションキーのみで回答してください (例: a0)。追加のテキストを含めないでください。\end{CJK}")

The text prompt template is the same as for Qwen2-VL and MiniCPM-V:

{
\footnotesize
\begin{verbatim}
text_prompt = (
        f"{question_str}\n"
        f"a0: {options['a0']}\n"
        f"a1: {options['a1']}\n"
        f"a2: {options['a2']}\n"
        f"a3: {options['a3']}\n"
        f"{instruction}"
    )
\end{verbatim}
}

The input contents can simply be the list [text\_prompt, video\_part], where the video with the proper MIME type, so that the model knows exactly what kind of data it takes. 

For all the models the answer is parsed with regex:
\begin{verbatim}
re.search(r"\ba[0-3]\b", generated_text.lower())
\end{verbatim}

\begin{sidewaystable*}[htp]
\centering
% Scale the table to fit the page height (in landscape orientation).
\resizebox{\textheight}{!}{%
\begin{tabular}{llllll}
\toprule
\textbf{Template} & \textbf{MCA} & \textbf{DCA} & \textbf{Condition} & \textbf{n\_persons} & \textbf{Question}\\
\midrule
1a & t & a & st\_mca >= et\_dca; st\_mca < et\_dca \& et\_mca > et\_dca & 1 & What did the person in the video do after v\_dca\_ing\_1 the obj\_1 adjunct conj v\_dca\_ing\_2 the obj\_2 ? \\
1b & t & a & st\_mca >= et\_dca; st\_mca < et\_dca \& et\_mca > et\_dca & 2 & What did the person in the video do after the other person person v\_dca\_past\_1 the obj\_1 adjunct conj v\_dca\_past\_2 the obj\_2 ? \\
2a1 & t & t & st\_mca >= et\_dca; st\_mca < et\_dca \& et\_mca > et\_dca & 1 & What did the person in the video do after they v\_dca\_past\_perf\_1 the obj\_1 adjunct conj v\_dca\_past\_perf\_2 the obj\_2 ? \\
2a2 & t & t & st\_mca >= et\_dca; st\_mca < et\_dca \& et\_mca > et\_dca & 1 & What did the person in the video do when they v\_dca\_past\_perf\_1 the obj\_1 adjunct conj v\_dca\_past\_perf\_2 the obj\_2 ? \\
2b1 & t & t & st\_mca >= et\_dca; st\_mca < et\_dca \& et\_mca > et\_dca & 2 & What did the person in the video do after the other person v\_dca\_past\_perf\_1 the obj\_1 adjunct conj v\_dca\_past\_perf\_2 the obj\_2 ? \\
2b2 & t & t & st\_mca >= et\_dca; st\_mca < et\_dca \& et\_mca > et\_dca & 2 & What did the person in the video do when the other person v\_dca\_past\_perf\_1 the obj\_1 adjunct conj v\_dca\_past\_perf\_2 the obj\_2 ? \\
3a & t & a & et\_mca <= st\_dca & 1 & What had the person in the video done before v\_dca\_ing\_1 the obj\_1 adjunct conj v\_dca\_ing\_2 the obj\_2 ? \\
3b & t & a & et\_mca <= st\_dca & 2 & What had the person in the video done before the other person v\_dca\_past\_1 the obj\_1 adjunct conj v\_dca\_past\_2 obj\_2 ? \\
4a & t & t & et\_mca <= st\_dca & 1 & What had the person in the video done before they v\_dca\_past\_1 the obj\_1 adjunct conj v\_dca\_past\_2 the obj\_2 ? \\
4b & t & t & et\_mca <= st\_dca & 2 & What had the person in the video done before the other person v\_dca\_past\_1 the obj\_1 adjunct conj v\_dca\_past\_2 the obj\_2 ? \\
5a1 & t & a & et\_mca > st\_dca \& et\_mca < et\_dca & 1 & What did the person in the video do v\_dca\_ing\_1 the obj\_1 adjunct conj v\_dca\_ing\_2 the obj\_2 ? \\
5a2 & t & a & et\_mca > st\_dca \& et\_mca < et\_dca & 1 & What did the person in the video do while v\_dca\_ing\_1 the obj\_1 adjunct conj v\_dca\_ing\_2 the obj\_2 ? \\
5b1 & t & a & et\_mca > st\_dca \& et\_mca < et\_dca & 2 & What did the person in the video do while the other person was v\_dca\_ing\_1 the obj\_1 adjunct conj v\_dca\_ing\_2 obj\_2 ? \\
5b2 & t & a & et\_mca > st\_dca \& et\_mca < et\_dca & 2 & What did the person in the video do when the other person was v\_dca\_ing\_1 the obj\_1 adjunct conj v\_dca\_ing\_2 the obj\_2 ? \\
6a & a & a & et\_mca <= st\_dca & 1 & What had the person in the video been doing before v\_dca\_ing\_1 the obj\_1 adjunct conj v\_dca\_ing\_2 the obj\_2 ? \\
6b & a & a & et\_mca <= st\_dca & 2 & What had the person in the video been doing before the other person was v\_dca\_ing\_1 the obj\_1 adjunct conj v\_dca\_past\_2 the obj\_2 ? \\
7a & a & t & et\_mca <= st\_dca & 1 & What had the person in the video been doing before v\_dca\_ing\_1 the obj\_1 adjunct conj v\_dca\_ing\_2 the obj\_2 ? \\
7b & a & t & et\_mca <= st\_dca & 2 & What had the person in the video been doing before the other person v\_dca\_past\_1 the obj\_1 adjunct conj v\_dca\_past\_2 the obj\_2 ? \\
8a1 & a & a & st\_mca > st\_dca \& st\_mca < et\_dca; st\_mca == st\_dca; st\_mca < st\_dca \& et\_mca > st\_dca & 1 & What was the person in the video doing while v\_dca\_ing\_1 the obj\_1 adjunct conj v\_dca\_ing\_2 the obj\_2 ? \\
8a2 & a & a & st\_mca > st\_dca \& st\_mca < et\_dca; st\_mca == st\_dca; st\_mca < st\_dca \& et\_mca > st\_dca & 1 & What was the person in the video doing when v\_dca\_ing\_1 the obj\_1 adjunct conj v\_dca\_ing\_2 the obj\_2 ? \\
8b1 & a & a & st\_mca > st\_dca \& st\_mca < et\_dca; st\_mca == st\_dca; st\_mca < st\_dca \& et\_mca > st\_dca & 2 & What was the person in the video doing while the other person was v\_dca\_ing\_1 the obj\_1 adjunct conj v\_dca\_ing\_2 the obj\_2 ? \\
8b2 & a & a & st\_mca > st\_dca \& st\_mca < et\_dca; st\_mca == st\_dca; st\_mca < st\_dca \& et\_mca > st\_dca & 2 & What was the person in the video doing when the other person was v\_dca\_ing\_1 the obj\_1 adjunct conj v\_dca\_ing\_2 the obj\_2 ? \\
9a & a & t & st\_mca < et\_dca \& et\_mca > et\_dca & 1 & What was the person in the video doing when they v\_dca\_past\_1 the obj\_1 adjunct conj v\_dca\_past\_2 the obj\_2 ? \\
9b & a & t & st\_mca < et\_dca \& et\_mca > et\_dca & 2 & What was the person in the video doing when the other person v\_dca\_past\_1 the obj\_1 adjunct conj v\_dca\_past\_2 the obj\_2 ? \\
10a & a & a & st\_mca >= et\_dca & 1 & What was the person in the video doing after v\_dca\_ing\_1 the obj\_1 adjunct conj v\_dca\_ing\_2 obj\_2 ? \\
10b & a & a & st\_mca >= et\_dca & 2 & What was the person in the video doing after the other person v\_dca\_past\_1 the obj\_1 adjunct conj v\_dca\_past\_2 the obj\_2 ? \\
11a & a & t & st\_mca >= et\_dca; st\_mca < et\_dca \& et\_mca > et\_dca & 1 & What was the person in the video doing after v\_dca\_ing\_1 the obj\_1 adjunct conj v\_dca\_ing\_2 the obj\_2 ? \\
11b & a & t & st\_mca >= et\_dca; st\_mca < et\_dca \& et\_mca > et\_dca & 2 & What was the person in the video doing after the other person v\_dca\_past\_1 the obj\_1 adjunct conj v\_dca\_past\_2 the obj\_2 ? \\
12a1 & t & t & et\_mca == et\_dca & 1 & What did the person in the video do when they v\_dca\_past\_1 the obj\_1 adjunct conj v\_dca\_past\_2 the obj\_2 ? \\
12a2 & t & t & et\_mca == et\_dca & 1 & What did the person in the video do the moment they v\_dca\_past\_1 the obj\_1 adjunct conj v\_dca\_past\_2 the obj\_2 ? \\
12b1 & t & t & et\_mca == et\_dca & 2 & What did the person in the video do when the other person v\_dca\_past\_1 the obj\_1 adjunct conj v\_dca\_past\_2 the obj\_2 ? \\
12b2 & t & t & et\_mca == et\_dca & 2 & What did the person in the video do the moment the other person v\_dca\_past\_1 the obj\_1 adjunct conj v\_dca\_past\_2 the obj\_2 ? \\
\bottomrule
\end{tabular}
}
\caption{English templates: \_past\_ is the past simple form, \_past\_perf\_ is the past perfect form, \_ing\_ is the -ing form including the past continuous form.}
\label{tab:vertical-acl-table}
\end{sidewaystable*}

\begin{sidewaystable*}[htp]
\centering
\resizebox{\textheight}{!}{%
\begin{tabular}{llllll}
\toprule
\textbf{Template} & \textbf{MCA} & \textbf{DCA} & \textbf{Condition} & \textbf{n\_persons} & \textbf{Question}\\
\midrule
1a & t & a & st\_mca >= et\_dca; st\_mca < et\_dca \& et\_mca > et\_dca & 1 & Cosa ha fatto la persona nel video dopo v\_inf\_pass\_pross\_1 obj\_1 adjunct conj v\_inf\_pass\_pross\_2 obj\_2 ? \\
1b & t & a & st\_mca >= et\_dca; st\_mca < et\_dca \& et\_mca > et\_dca & 2 & Cosa ha fatto la persona nel video dopo che l'altra persona v\_trapass\_pross\_1 obj\_1 adjunct conj v\_trapass\_pross\_2 obj\_2 ? \\
2a1 & t & t & st\_mca >= et\_dca; st\_mca < et\_dca \& et\_mca > et\_dca & 1 & Cosa fece la persona nel video dopo v\_inf\_pass\_pross\_1 obj\_1 adjunct conj v\_inf\_pass\_pross\_2 obj\_2 ? \\
2a2 & t & t & st\_mca >= et\_dca; st\_mca < et\_dca \& et\_mca > et\_dca & 1 & Cosa fece la persona nel video quando v\_pass\_rem\_1 obj\_1 adjunct conj v\_pass\_rem\_2 obj\_2 ? \\
2b1 & t & t & st\_mca >= et\_dca; st\_mca < et\_dca \& et\_mca > et\_dca & 2 & Cosa fece la persona nel video dopo che l'altra persona v\_pass\_rem\_1 obj\_1 adjunct conj v\_pass\_rem\_2 obj\_2 ? \\
2b2 & t & t & st\_mca >= et\_dca; st\_mca < et\_dca \& et\_mca > et\_dca & 2 & Cosa fece la persona nel video quando l'altra persona v\_pass\_rem\_1 obj\_1 adjunct conj v\_pass\_rem\_2 obj\_2 ? \\
3a & t & a & et\_mca <= st\_dca & 1 & Cosa aveva fatto la persona nel video prima di v\_inf\_1 obj\_1 adjunct conj v\_inf\_2 obj\_2 ? \\
3b & t & a & et\_mca <= st\_dca & 2 & Cosa aveva fatto la persona nel video prima che l'altra persona v\_imp\_1 obj\_1 adjunct conj v\_imp\_2 obj\_2 ? \\
4a & t & t & et\_mca <= st\_dca & 1 & Cosa aveva fatto la persona nel video prima di v\_inf\_1 obj\_1 adjunct conj v\_inf\_2 obj\_2 ? \\
4b & t & t & et\_mca <= st\_dca & 2 & Cosa aveva fatto la persona nel video prima che l'altra persona v\_trapass\_cong\_1 obj\_1 adjunct conj v\_trapass\_cong\_2 obj\_2 ? \\
5a1 & t & a & et\_mca > st\_dca \& et\_mca < et\_dca & 1 & Cosa ha fatto la persona nel video v\_ger\_1 obj\_1 adjunct conj v\_ger\_2 obj\_2 ? \\
5a2 & t & a & et\_mca > st\_dca \& et\_mca < et\_dca & 1 & Cosa ha fatto la persona nel video mentre v\_imp\_1 obj\_1 adjunct conj v\_imp\_2 obj\_2 ? \\
5b1 & t & a & et\_mca > st\_dca \& et\_mca < et\_dca & 2 & Cosa ha fatto la persona nel video mentre l'altra persona v\_imp\_1 obj\_1 adjunct conj v\_imp\_2 obj\_2 ? \\
5b2 & t & a & et\_mca > st\_dca \& et\_mca < et\_dca & 2 & Cosa ha fatto la persona nel video quando l'altra persona v\_imp\_1 obj\_1 adjunct conj v\_imp\_2 obj\_2 ? \\
6a & a & a & et\_mca <= st\_dca & 1 & Cosa stava facendo la persona nel video prima di v\_inf\_1 obj\_1 adjunct conj v\_inf\_2 obj\_2 ? \\
6b & a & a & et\_mca <= st\_dca & 2 & Cosa stava facendo la persona nel video prima che l'altra persona v\_imp\_cong\_1 obj\_1 adjunct conj v\_imp\_cong\_2 obj\_2 ? \\
7a & a & t & et\_mca <= st\_dca & 1 & Cosa stava facendo la persona nel video prima di v\_inf\_1 obj\_1 adjunct conj v\_inf\_2 obj\_2 ? \\
7b & a & t & et\_mca <= st\_dca & 2 & Cosa stava facendo la persona nel video prima che l'altra persona v\_imp\_cong\_1 obj\_1 adjunct conj v\_inf\_2 obj\_2 ? \\
8a1 & a & a & st\_mca > st\_dca \& st\_mca < et\_dca; st\_mca == st\_dca; st\_mca < st\_dca \& et\_mca > st\_dca & 1 & Cosa faceva la persona nel video mentre v\_imp\_prog\_1 obj\_1 adjunct conj v\_imp\_prog\_2 obj\_2 ? \\
8a2 & a & a & st\_mca > st\_dca \& st\_mca < et\_dca; st\_mca == st\_dca; st\_mca < st\_dca \& et\_mca > st\_dca & 1 & Cosa faceva la persona nel video quando v\_imp\_prog\_1 obj\_1 adjunct conj v\_imp\_prog\_2 obj\_2 ? \\
8b1 & a & a & st\_mca > st\_dca \& st\_mca < et\_dca; st\_mca == st\_dca; st\_mca < st\_dca \& et\_mca > st\_dca & 2 & Cosa faceva la persona nel video mentre l'altra persona v\_imp\_prog\_1 obj\_1 adjunct conj v\_imp\_prog\_2 obj\_2 ? \\
8b2 & a & a & st\_mca > st\_dca \& st\_mca < et\_dca; st\_mca == st\_dca; st\_mca < st\_dca \& et\_mca > st\_dca & 2 & Cosa faceva la persona nel video quando l'altra persona v\_imp\_prog\_1 obj\_1 adjunct conj v\_imp\_prog\_2 obj\_2 ? \\9a & a & t & st\_mca < et\_dca \& et\_mca > et\_dca & 1 & Cosa faceva la persona nel video quando v\_imp\_1 obj\_1 adjunct conj v\_imp\_2 obj\_2 ? \\
9b & a & t & st\_mca < et\_dca \& et\_mca > et\_dca & 2 & Cosa faceva la persona nel video quando l'altra persona v\_imp\_1 obj\_1 adjunct conj v\_imp\_2 obj\_2 ? \\
10a & a & a & st\_mca >= et\_dca & 1 & Cosa faceva la persona nel video dopo v\_inf\_pass\_pross\_1 obj\_1 adjunct conj v\_inf\_pass\_pross\_2 obj\_2 ? \\
10b & a & a & st\_mca >= et\_dca & 2 & Cosa faceva la persona nel video dopo che l'altra persona v\_trapass\_pross\_1 obj\_1 adjunct conj v\_trapass\_pross\_2 obj\_2 ? \\
12a1 & t & t & et\_mca == et\_dca & 1 & Cosa fece la persona nel video quando v\_pass\_rem\_1 obj\_1 adjunct conj v\_pass\_rem\_2 obj\_2 ? \\
12a2 & t & t & et\_mca == et\_dca & 1 & Cosa fece la persona nel video nel momento in cui v\_pass\_rem\_1 obj\_1 adjunct conj v\_pass\_rem\_2 obj\_2 ? \\
12b1 & t & t & et\_mca == et\_dca & 2 & Cosa fece la persona nel video quando l'altra persona v\_pass\_rem\_1 obj\_1 adjunct conj v\_pass\_rem\_2 obj\_2 ? \\
12b2 & t & t & et\_mca == et\_dca & 2 & Cosa fece la persona nel video nel momento in cui l'altra persona v\_pass\_rem\_1 obj\_1 adjunct conj v\_pass\_rem\_2 obj\_2 ? \\
\bottomrule
\end{tabular}
}
\caption{Italian templates: \_inf\_ is the infinitive form, \_pass\_pross\_ is the passato prossimo (present perfect) form, \_trapass\_pross\_ is the trapassato prossimo (past perfect) form, \_pass\_rem\_ is the passato remoto (simple past) form, \_imp\_ is the imperfetto (imperfect) form, \_ger\_ is the gerundio (gerund) form, \_imp\_cong\_ is the imperfetto congiuntivo (imperfect subjunctive) form, \_imp\_prog\_ is the imperfetto progressivo (past continuous) form. Conj is a coordinating conjunction in case of several verbs in one class, as in \textit{lavorando o giocando al computer}.}
\label{tab:italian-templates}
\end{sidewaystable*}

\begin{sidewaystable*}[htp]
\centering
\resizebox{\textheight}{!}{%
\begin{tabular}{llllll}
\toprule
\textbf{Template} & \textbf{MCA} & \textbf{DCA} & \textbf{Condition} & \textbf{n\_persons} & \textbf{Question} \\
\midrule
1a  & t & a & st\_mca >= et\_dca; st\_mca < et\_dca \& et\_mca > et\_dca & 1 & \selectlanguage{russian}Что сделал человек на видео после того, как\selectlanguage{english} v\_imp\_1 obj\_1 adjunct conj v\_imp\_2 obj\_2 ? \\
1b  & t & a & st\_mca >= et\_dca; st\_mca < et\_dca \& et\_mca > et\_dca & 2 & \selectlanguage{russian}Что сделал человек на видео после того, как другой человек\selectlanguage{english} v\_imp\_1 obj\_1 adjunct conj v\_imp\_2 obj\_2 ? \\
2a1 & t & t & st\_mca >= et\_dca; st\_mca < et\_dca \& et\_mca > et\_dca & 1 & \selectlanguage{russian}Что сделал человек на видео после того, как\selectlanguage{english} v\_perf\_1 obj\_1 adjunct conj v\_perf\_2 obj\_2 ? \\
2a2 & t & t & st\_mca >= et\_dca; st\_mca < et\_dca \& et\_mca > et\_dca & 1 & \selectlanguage{russian}Что сделал человек на видео, когда\selectlanguage{english} v\_perf\_1 obj\_1 adjunct conj v\_perf\_2 obj\_2 ? \\
2b1 & t & t & st\_mca >= et\_dca; st\_mca < et\_dca \& et\_mca > et\_dca & 2 & \selectlanguage{russian}Что сделал человек на видео после того, как другой человек\selectlanguage{english} v\_perf\_1 obj\_1 adjunct conj v\_perf\_2 obj\_2 ? \\
2b2 & t & t & st\_mca >= et\_dca; st\_mca < et\_dca \& et\_mca > et\_dca & 2 & \selectlanguage{russian}Что сделал человек на видео, когда другой человек\selectlanguage{english} v\_perf\_1 obj\_1 adjunct conj v\_perf\_2 obj\_2 ? \\
3a  & t & a & et\_mca <= st\_dca & 1 & \selectlanguage{russian}Что сделал человек на видео до того, как\selectlanguage{english} v\_imp\_1 obj\_1 adjunct conj v\_imp\_2 obj\_2 ? \\
3b  & t & a & et\_mca <= st\_dca & 2 & \selectlanguage{russian}Что сделал человек на видео перед тем, как другой человек\selectlanguage{english} v\_imp\_1 obj\_1 adjunct conj v\_imp\_2 obj\_2 ? \\
4a  & t & t & et\_mca <= st\_dca & 1 & \selectlanguage{russian}Что сделал человек на видео до того, как\selectlanguage{english} v\_perf\_1 obj\_1 adjunct conj v\_perf\_2 obj\_2 ? \\
4b  & t & t & et\_mca <= st\_dca & 2 & \selectlanguage{russian}Что сделал человек перед тем, как другой человек\selectlanguage{english} v\_perf\_1 obj\_1 adjunct conj v\_perf\_2 obj\_2 ? \\
5a1 & t & a & et\_mca > st\_dca \& et\_mca < et\_dca & 1 & \selectlanguage{russian}Что сделал человек, пока он\selectlanguage{english} v\_imp\_1 obj\_1 adjunct conj v\_imp\_2 obj\_2 ? \\
5a2 & t & a & et\_mca > st\_dca \& et\_mca < et\_dca & 1 & \selectlanguage{russian}Что сделал человек, когда\selectlanguage{english} v\_imp\_1 obj\_1 adjunct conj v\_imp\_2 obj\_2 ? \\
5b1 & t & a & et\_mca > st\_dca \& et\_mca < et\_dca & 2 & \selectlanguage{russian}Что сделал человек, пока другой человек\selectlanguage{english} v\_imp\_1 obj\_1 adjunct conj v\_imp\_2 obj\_2 ? \\
5b2 & t & a & et\_mca > st\_dca \& et\_mca < et\_dca & 2 & \selectlanguage{russian}Что сделал человек, когда другой человек\selectlanguage{english} v\_imp\_1 obj\_1 adjunct conj v\_imp\_2 obj\_2 ? \\
6a  & a & a & et\_mca <= st\_dca & 1 & \selectlanguage{russian}Что делал человек до того, как он\selectlanguage{english} v\_imp\_1 obj\_1 adjunct conj v\_imp\_2 obj\_2 ? \\
6b  & a & a & et\_mca <= st\_dca & 2 & \selectlanguage{russian}Что делал человек до того, как другой человек\selectlanguage{english} v\_imp\_1 obj\_1 adjunct conj v\_imp\_2 obj\_2 ? \\
7a  & a & t & et\_mca <= st\_dca & 1 & \selectlanguage{russian}Что делал человек до того, как он\selectlanguage{english} v\_perf\_1 obj\_1 adjunct conj v\_perf\_2 obj\_2 ? \\
7b  & a & t & et\_mca <= st\_dca & 2 & \selectlanguage{russian}Что делал человек до того, как другой человек\selectlanguage{english} v\_perf\_1 obj\_1 adjunct conj v\_perf\_2 obj\_2 ? \\
8a1 & a & a & st\_mca > st\_dca \& st\_mca < et\_dca; st\_mca == st\_dca; st\_mca < st\_dca \& et\_mca > st\_dca & 1 & \selectlanguage{russian}Что делал человек, пока\selectlanguage{english} v\_imp\_1 obj\_1 adjunct conj v\_imp\_2 obj\_2 ? \\
8a2 & a & a & st\_mca > st\_dca \& st\_mca < et\_dca; st\_mca == st\_dca; st\_mca < st\_dca \& et\_mca > st\_dca & 1 & \selectlanguage{russian}Что делал человек, когда\selectlanguage{english} v\_imp\_1 obj\_1 adjunct conj v\_imp\_2 obj\_2 ? \\
8b1 & a & a & st\_mca > st\_dca \& st\_mca < et\_dca; st\_mca == st\_dca; st\_mca < st\_dca \& et\_mca > st\_dca & 2 & \selectlanguage{russian}Что делал человек, пока другой человек\selectlanguage{english} v\_imp\_1 obj\_1 adjunct conj v\_imp\_2 obj\_2 ? \\
8b2 & a & a & st\_mca > st\_dca \& st\_mca < et\_dca; st\_mca == st\_dca; st\_mca < st\_dca \& et\_mca > st\_dca & 2 & \selectlanguage{russian}Что делал человек, когда другой человек\selectlanguage{english} v\_imp\_1 obj\_1 adjunct conj v\_imp\_2 obj\_2 ? \\
9a  & a & t & st\_mca < et\_dca \& et\_mca > et\_dca & 1 & \selectlanguage{russian}Что делал человек, когда он\selectlanguage{english} v\_perf\_1 obj\_1 adjunct conj v\_perf\_2 obj\_2 ? \\
9b  & a & t & st\_mca < et\_dca \& et\_mca > et\_dca & 2 & \selectlanguage{russian}Что делал человек, когда другой человек\selectlanguage{english} v\_perf\_1 obj\_1 adjunct conj v\_perf\_2 obj\_2 ? \\
10a  & a & a & st\_mca >= et\_dca & 1 & \selectlanguage{russian}Что делал человек после того, как он\selectlanguage{english} v\_imp\_1 obj\_1 adjunct conj v\_imp\_2 obj\_2 ? \\
10b  & a & a & st\_mca >= et\_dca & 2 & \selectlanguage{russian}Что делал человек после того, как другой человек\selectlanguage{english} v\_imp\_1 obj\_1 adjunct conj v\_imp\_2 obj\_2 ? \\
11a  & a & t & st\_mca >= et\_dca; st\_mca < et\_dca \& et\_mca > et\_dca & 1 & \selectlanguage{russian}Что делал человек после того, как он\selectlanguage{english} v\_perf\_1 obj\_1 adjunct conj v\_perf\_2 obj\_2 ? \\
11b  & a & t & st\_mca >= et\_dca; st\_mca < et\_dca \& et\_mca > et\_dca & 2 & \selectlanguage{russian}Что делал человек после того, как другой человек\selectlanguage{english} v\_perf\_1 obj\_1 adjunct conj v\_perf\_2 obj\_2 ? \\
12a1 & t & t & et\_mca == et\_dca & 1 & \selectlanguage{russian}Что сделал человек на видео, когда\selectlanguage{english} v\_perf\_1 obj\_1 adjunct conj v\_perf\_2 obj\_2 ? \\
12a2 & t & t & et\_mca == et\_dca & 1 & \selectlanguage{russian}Что сделал человек на видео в тот момент, когда\selectlanguage{english} v\_perf\_1 obj\_1 adjunct conj v\_perf\_2 obj\_2 ? \\
12b1 & t & t & et\_mca == et\_dca & 2 & \selectlanguage{russian}Что сделал человек на видео, когда другой человек\selectlanguage{english} v\_perf\_1 obj\_1 adjunct conj v\_perf\_2 obj\_2 ? \\
12b2 & t & t & et\_mca == et\_dca & 2 & \selectlanguage{russian}Что сделал человек на видео в тот момент, когда другой человек\selectlanguage{english} v\_perf\_1 obj\_1 adjunct conj v\_perf\_2 obj\_2 ? \\
\bottomrule
\end{tabular}%
}
\caption{Russain Templates: \_imp\_ is imperfective past, \_perf\_ — perfective past. Conj is a coordinating conjunction in case of several verbs in one class, as in \textit{\selectlanguage{russian}читал или писал\selectlanguage{english}}.}
\label{tab:russian-templates}
\end{sidewaystable*}

\begin{sidewaystable*}[htp]
\centering
\resizebox{\textheight}{!}{
\begin{tabular}{llllll}
\toprule
\textbf{Template} & \textbf{MCA} & \textbf{DCA} & \textbf{Condition} & \textbf{n\_persons} & \textbf{Question} \\
\midrule
1a & t & a & st\_mca >= et\_dca; st\_mca < et\_dca \& et\_mca > et\_dca & 1 & \begin{CJK}{UTF8}{min}ビデオに写っている人は\end{CJK}adj obj\begin{CJK}{UTF8}{min}を\end{CJK}ta\_form \begin{CJK}{UTF8}{min}後、何をしましたか？\end{CJK} \\
1b  & t & a & st\_mca >= et\_dca; st\_mca < et\_dca \& et\_mca > et\_dca & 2 & \begin{CJK}{UTF8}{min}ビデオに写っている人は、他の人が\end{CJK}adj obj\begin{CJK}{UTF8}{min}を\end{CJK}ta\_form \begin{CJK}{UTF8}{min}後、何をしましたか？\end{CJK} \\
2a1 & t & t & st\_mca >= et\_dca; st\_mca < et\_dca \& et\_mca > et\_dca & 1 & \begin{CJK}{UTF8}{min}ビデオに写っている人は\end{CJK}adj obj\begin{CJK}{UTF8}{min}を\end{CJK}ta\_form \begin{CJK}{UTF8}{min}後、何をしましたか？\end{CJK} \\
2a2 & t & t & st\_mca >= et\_dca; st\_mca < et\_dca \& et\_mca > et\_dca & 1 & \begin{CJK}{UTF8}{min}ビデオに写っている人は\end{CJK}adj obj\begin{CJK}{UTF8}{min}を\end{CJK}ta\_form \begin{CJK}{UTF8}{min}時、何をしましたか？\end{CJK} \\
2b1 & t & t & st\_mca >= et\_dca; st\_mca < et\_dca \& et\_mca > et\_dca & 2 & \begin{CJK}{UTF8}{min}ビデオに写っている人は、他の人が\end{CJK}adj obj\begin{CJK}{UTF8}{min}を\end{CJK}ta\_form \begin{CJK}{UTF8}{min}後、何をしましたか？\end{CJK} \\
2b2 & t & t & st\_mca >= et\_dca; st\_mca < et\_dca \& et\_mca > et\_dca & 2 & \begin{CJK}{UTF8}{min}ビデオに写っている人は、他の人が\end{CJK}adj obj\begin{CJK}{UTF8}{min}を\end{CJK}ta\_form \begin{CJK}{UTF8}{min}時、何をしましたか？\end{CJK} \\
3a & t & a & et\_mca <= st\_dca & 1 & \begin{CJK}{UTF8}{min}ビデオに写っている人は\end{CJK}adj obj\begin{CJK}{UTF8}{min}を\end{CJK}inf\begin{CJK}{UTF8}{min}前に、何をしましたか？\end{CJK} \\
3b  & t & a & et\_mca <= st\_dca & 2 & \begin{CJK}{UTF8}{min}ビデオに写っている人は、他の人が\end{CJK}adj obj\begin{CJK}{UTF8}{min}を\end{CJK}inf\begin{CJK}{UTF8}{min}前に、何をしましたか？\end{CJK} \\
4a  & t & t & et\_mca <= st\_dca & 1 & \begin{CJK}{UTF8}{min}ビデオに写っている人は\end{CJK}adj obj\begin{CJK}{UTF8}{min}を\end{CJK}inf\begin{CJK}{UTF8}{min}前に、何をしましたか？\end{CJK} \\
4b  & t & t & et\_mca <= st\_dca & 2 & \begin{CJK}{UTF8}{min}ビデオに写っている人は、他の人が\end{CJK}adj obj\begin{CJK}{UTF8}{min}を\end{CJK}inf\begin{CJK}{UTF8}{min}前に、何をしましたか？\end{CJK} \\
5a1 & t & a & et\_mca > st\_dca \& et\_mca < et\_dca & 1 & \begin{CJK}{UTF8}{min}ビデオに写っている人は\end{CJK}adj obj\begin{CJK}{UTF8}{min}を\end{CJK}te\_form \begin{CJK}{UTF8}{min}、何をしましたか？\end{CJK} \\
5a2 & t & a & et\_mca > st\_dca \& et\_mca < et\_dca & 1 & \begin{CJK}{UTF8}{min}ビデオに写っている人は\end{CJK}adj obj\begin{CJK}{UTF8}{min}を\end{CJK}i\_form \begin{CJK}{UTF8}{min}ながら、何をしましたか？\end{CJK} \\
5b1 & t & a & et\_mca > st\_dca \& et\_mca < et\_dca & 2 & \begin{CJK}{UTF8}{min}ビデオに写っている人は、他の人が\end{CJK}adj obj\begin{CJK}{UTF8}{min}を\end{CJK}inf\_imp\begin{CJK}{UTF8}{min}間に、何をしましたか？\end{CJK} \\
5b2 & t & a & et\_mca > st\_dca \& et\_mca < et\_dca & 2 & \begin{CJK}{UTF8}{min}ビデオに写っている人は、他の人が\end{CJK}adj obj\begin{CJK}{UTF8}{min}を\end{CJK}ta\_form \begin{CJK}{UTF8}{min}時、何をしましたか？\end{CJK} \\
6a  & a & a & et\_mca <= st\_dca & 1 & \begin{CJK}{UTF8}{min}ビデオに写っている人は\end{CJK}adj obj\begin{CJK}{UTF8}{min}を\end{CJK}inf\begin{CJK}{UTF8}{min}前に、何をしていましたか？\end{CJK} \\
6b  & a & a & et\_mca <= st\_dca & 2 & \begin{CJK}{UTF8}{min}ビデオに写っている人は、他の人が\end{CJK}adj obj\begin{CJK}{UTF8}{min}を\end{CJK}inf\_imp\begin{CJK}{UTF8}{min}前に、何をしていましたか？\end{CJK} \\
7a  & a & t & et\_mca <= st\_dca & 1 & \begin{CJK}{UTF8}{min}ビデオに写っている人は\end{CJK}adj obj\begin{CJK}{UTF8}{min}を\end{CJK}inf\begin{CJK}{UTF8}{min}前に、何をしていましたか？\end{CJK} \\
7b  & a & t & et\_mca <= st\_dca & 2 & \begin{CJK}{UTF8}{min}ビデオに写っている人は、他の人が\end{CJK}adj obj\begin{CJK}{UTF8}{min}を\end{CJK}inf\begin{CJK}{UTF8}{min}前に、何をしていましたか？\end{CJK} \\
8a1 & a & a & st\_mca > st\_dca \& st\_mca < et\_dca; st\_mca = st\_dca; st\_mca < st\_dca \& et\_mca > st\_dca & 1 & \begin{CJK}{UTF8}{min}ビデオに写っている人は\end{CJK}adj obj\begin{CJK}{UTF8}{min}を\end{CJK}i\_form \begin{CJK}{UTF8}{min}ながら、何をしていましたか？\end{CJK} \\
8a2 & a & a & st\_mca > st\_dca \& st\_mca < et\_dca; st\_mca = st\_dca; st\_mca < st\_dca \& et\_mca > st\_dca & 1 & \begin{CJK}{UTF8}{min}ビデオに写っている人は\end{CJK}adj obj\begin{CJK}{UTF8}{min}を\end{CJK}ta\_form \begin{CJK}{UTF8}{min}時、何をしていましたか？\end{CJK} \\
8b1 & a & a & st\_mca > st\_dca \& st\_mca < et\_dca; st\_mca = st\_dca; st\_mca < st\_dca \& et\_mca > st\_dca & 2 & \begin{CJK}{UTF8}{min}ビデオに写っている人は、他の人がadj objをinf\_imp間に、何をしていましたか？\end{CJK} \\
8b2 & a & a & st\_mca > st\_dca \& st\_mca < et\_dca; st\_mca = st\_dca; st\_mca < st\_dca \& et\_mca > st\_dca & 2 & \begin{CJK}{UTF8}{min}ビデオに写っている人は、他の人がadj objをta\_form時、何をしていましたか？\end{CJK} \\
9a  & a & t & st\_mca < et\_dca \& et\_mca > et\_dca & 1 & \begin{CJK}{UTF8}{min}ビデオに写っている人はadj objをta\_form時、何をしていましたか？\end{CJK} \\
9b  & a & t & st\_mca < et\_dca \& et\_mca > et\_dca & 2 & \begin{CJK}{UTF8}{min}ビデオに写っている人は、他の人がadj objをta\_form時、何をしていましたか？\end{CJK} \\
10a  & a & a & st\_mca >= et\_dca & 1 & \begin{CJK}{UTF8}{min}ビデオに写っている人はadj objをimp後、何をしていましたか？\end{CJK} \\
10b  & a & a & st\_mca >= et\_dca & 2 & \begin{CJK}{UTF8}{min}ビデオに写っている人は、他の人がadj objをimp後、何をしていましたか？\end{CJK} \\
11a  & a & t & st\_mca >= et\_dca; st\_mca < et\_dca \& et\_mca > et\_dca & 1 & \begin{CJK}{UTF8}{min}ビデオに写っている人はadj objをta\_form後、何をしていましたか？\end{CJK} \\
11b  & a & t & st\_mca >= et\_dca; st\_mca < et\_dca \& et\_mca > et\_dca & 2 & \begin{CJK}{UTF8}{min}ビデオに写っている人は、他の人がadj objをta\_form後、何をしていましたか？\end{CJK} \\
12a1 & t & t & et\_mca = et\_dca & 1 & \begin{CJK}{UTF8}{min}ビデオに写っている人はadj objをta\_form時、何をしましたか？\end{CJK} \\
12a2 & t & t & et\_mca = et\_dca & 1 & \begin{CJK}{UTF8}{min}ビデオに写っている人はadj objをta\_form瞬間、何をしましたか？\end{CJK} \\
12b1 & t & t & et\_mca = et\_dca & 2 & \begin{CJK}{UTF8}{min}ビデオに写っている人は、他の人がadj objをta\_form時、何をしましたか？\end{CJK} \\
12b2 & t & t & et\_mca = et\_dca & 2 & \begin{CJK}{UTF8}{min}ビデオに写っている人は、他の人が\end{CJK}adj obj\begin{CJK}{UTF8}{min}を\end{CJK}ta\_form \begin{CJK}{UTF8}{min}瞬間、何をしましたか？\end{CJK} \\
\bottomrule
\end{tabular}
}
\caption{Japanese templates: ta\_form is the general past form, inf is the dictionary form, te\_form is the conjunction form, i\_form is the present progressive, and imp is the imperfect. Conj is a coordinating conjunction in case of several verbs in one class, as in \begin{CJK}{UTF8}{min}ノートパソコンで作業したり遊んだりする\end{CJK}.}
\label{tab:japanese-templates}
\end{sidewaystable*}

\begin{table*}[ht]
\centering
\small
\setlength{\tabcolsep}{8pt}
\renewcommand{\arraystretch}{1.2}
\caption{Annotator accuracy and distractor type distribution.}
\label{tab:annotator_stats}
\resizebox{\textwidth}{!}{%
\begin{tabular}{lcccc}
\toprule
\textbf{Annotator} & \textbf{Accuracy} & \textbf{Distractor Type 1} & \textbf{Distractor Type 2} & \textbf{Distractor Type 3} \\
\midrule
English  & 93.40 & 45.62 & 50.88 & 3.50 \\
Italian  & 97.84 & 20.83 & 12.50 & 66.67 \\
Russian  & 83.83 & 34.57 & 46.91 & 18.52 \\
Japanese & 98.36 & 20.00 & 40.00 & 40.00 \\
\bottomrule
\end{tabular}%
}
\end{table*}

\begin{table*}[h!]
\centering
\small
\setlength{\tabcolsep}{8pt}
\renewcommand{\arraystretch}{1.2}
\caption{Model configurations and references.}
\label{tab:model_configs}
\resizebox{\textwidth}{!}{%
\begin{tabular}{lccc}
\toprule
\textbf{Model} & \textbf{Vision Encoder} & \textbf{Language Model} & \textbf{Parameters} \\
\midrule
LLaVA-NeXT-Video-7B  & SigLIP-400M \cite{zhai2023sigmoid} & Qwen 1.5 \cite{qwen} & 7B \\
MiniCPM-V-2\_6       & SigLIP-400M \cite{zhai2023sigmoid} & Qwen2 \cite{qwen}    & 7B \\
Qwen2-VL             & Qwen2-VL \cite{Qwen2VL}            & Qwen2 \cite{qwen}    & 7B \\
InternVL2            & InternViT-300M-448px \cite{chen2024internvl} & internlm2\_5-7b-chat \cite{chen2024internvl} & 8B \\
\bottomrule
\end{tabular}%
}
\end{table*}

\begin{table*}[t]
\caption{The proportion of correctly predicted answers with respect to telicity in percentage.}
\label{tab:telicity_accuracy}
\centering
\small
\begin{tabular}{lcccc}
\toprule
\textbf{} & \textbf{Correct Telic} & \textbf{Incorrect Telic} & \textbf{Correct Atelic} & \textbf{Incorrect Atelic} \\
\midrule
Gemini-2.0-flash-lite & 42.97 & 57.03 & 41.97 & 58.03 \\
Qwen2-VL-7B-Instruct  & 42.80 & 57.20 & 32.95 & 67.05 \\
MiniCPM-V-2\_6        & 41.40 & 58.60 & 33.75 & 66.25 \\
GPT-4o                & 39.74 & 60.26 & 43.45 & 56.55 \\
InternVL2-8B          & 34.26 & 65.74 & 29.82 & 70.18 \\
LLaVA-NeXT-Video-7B   & 28.03 & 71.97 & 27.62 & 72.38 \\
\bottomrule
\end{tabular}
\end{table*}

\begin{table}[t]
\caption{Number of questions answered correctly by all models for each language. The 5 Models Column shows the results for all models except LLaVA-NeXT-Video-7B.}
\label{tab:all_correct_lang}
\centering
\small
\begin{tabular}{lcc}
\toprule
\textbf{Language} & \textbf{5 Models} & \textbf{All Models} \\
\midrule
English  & 230 & 81 \\
Italian  & 249 & 14 \\
Russian  & 211 & 49 \\
Japanese & 154 & 26 \\
\bottomrule
\end{tabular}
\end{table}

\begin{table}[t]
\caption{Number of questions answered incorrectly by all models for each language. The 5 Models Column shows the results for all models except LLaVA-NeXT-Video-7B.}
\label{tab:all_incorrect_lang}
\centering
\small
\begin{tabular}{lcc}
\toprule
\textbf{Language} & \textbf{5 Models} & \textbf{All Models} \\
\midrule
English  & 857 & 608 \\
Italian  & 808 & 456 \\
Russian  & 740 & 512 \\
Japanese & 850 & 554 \\
\bottomrule
\end{tabular}
\end{table}

\section{Statistics on Correct and Incorrect Answers}
\label{sec:corr}

Table \ref{tab:telicity_accuracy} shows the ratio of telic and atelic classes in the results of each model. With the exception of GPT-4o, all models perform better on questions about telic classes.

Tables \ref{tab:all_correct_lang}-\ref{tab:all_incorrect_lang} gives the intersection of correctly and incorrectly answered questions by all models respectively. As  LLaVA-NeXT-
Video-7B performed poorly on languages other than English, we give the results with and without it. 

\section{Statistics on Conjunctions}
\label{sec:conj}

\begin{figure}[ht]
\centering
\includegraphics[width=\columnwidth]{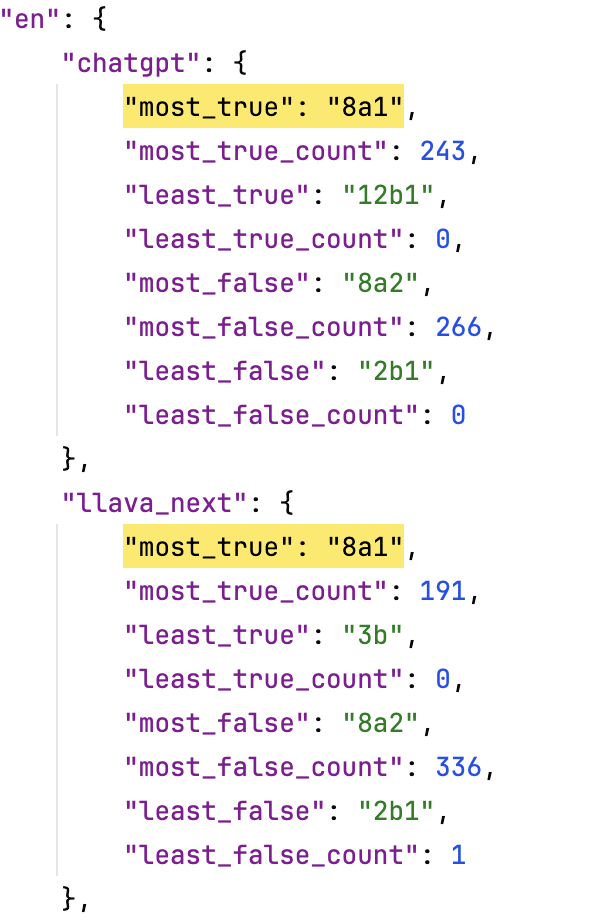}
\caption{Statistics on while (8a1) vs. when (8a2) for English in GPT-4o and LLaVA-NeXT-Video.}
\label{fig:eng_gpt}
\end{figure}

\begin{figure}[!h]
\centering
\includegraphics[width=\columnwidth]{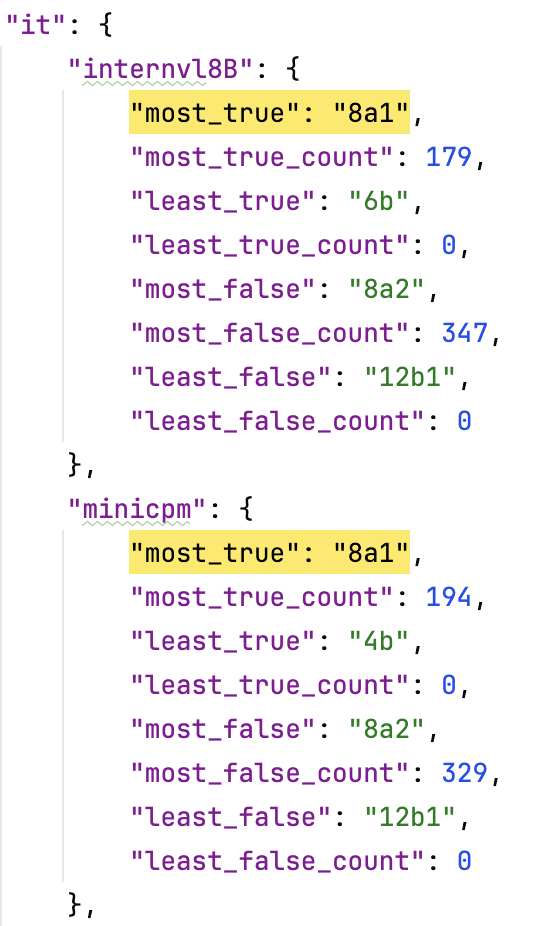}
\caption{Statistics on \textit{mentre} (8a1) vs. \textit{quando} (8a2) for Italian in InternVL2 and MiniCPM-V.}
\label{fig:it_int}
\end{figure}

\begin{figure}[!h]
\centering
\includegraphics[width=\columnwidth]{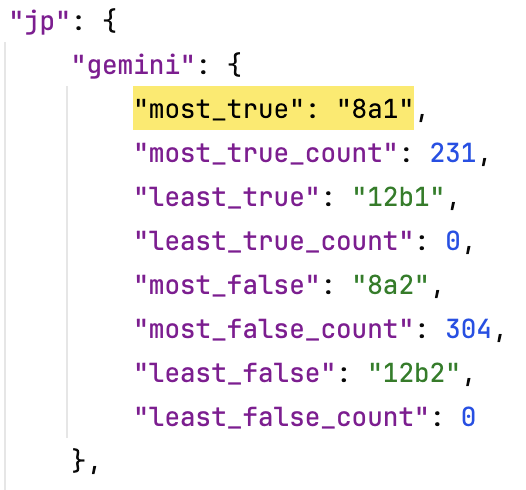}
\caption{Statistics on \begin{CJK}{UTF8}{min}ながら\end{CJK} (8a1) vs. \begin{CJK}{UTF8}{min}時\end{CJK} (8a2) for Japanese in Gemini-2.0-Flash-Lite.}
\label{fig:jp_gem}
\end{figure}

\begin{figure}[ht]
\centering
\includegraphics[width=\columnwidth]{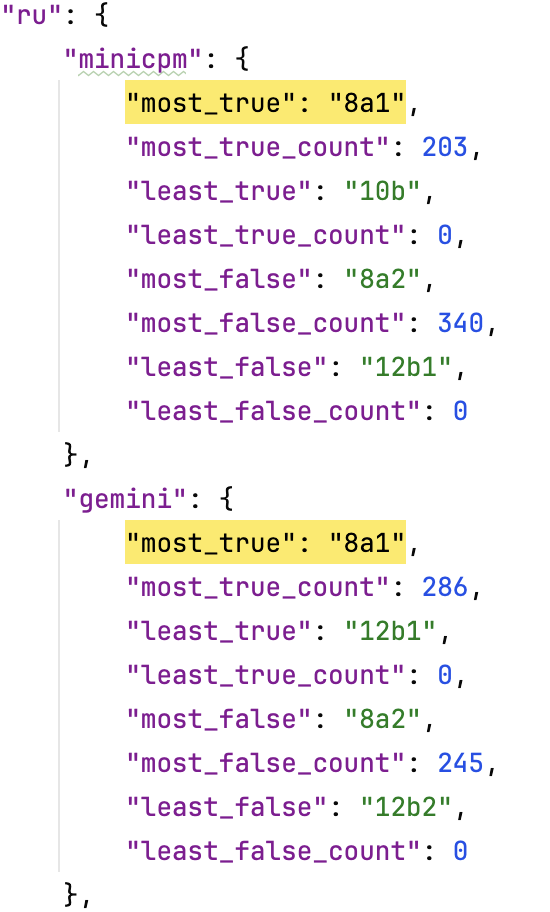}
\caption{Statistics on \selectlanguage{russian}пока\selectlanguage{english} (8a1) vs. \selectlanguage{russian}когда\selectlanguage{english} (8a2) for Russian in MiniCPM-V and Gemini-2.0-Flash-Lite.}
\label{fig:ru_mini}
\end{figure}

Figures \ref{fig:eng_gpt}-\ref{fig:ru_mini} show preference of the correct answers with \textit{while} over the ones with \textit{when} across all the languages in numbers. 

Tables \ref{tab:top5_templates}-\ref{tab:worst5_model_templates} show Top-5 Best/Worst Answered questions by templates corresponding to the sequential and durative conjunctions. 

\begin{table}[t]
\centering
\caption{Top-5 best answered templates by language across all models. Templates for questions about the following event are in \textcolor{blue}{blue}, templates for questions about preceding events in \textcolor{red}{red}, and templates for questions about simultaneous events in \textcolor{green}{green}.}
\label{tab:top5_templates}
\small
\begin{tabular}{lccccc}
\toprule
\textbf{} & \textbf{1} & \textbf{2} & \textbf{3} & \textbf{4} & \textbf{5} \\
\midrule
English  & \textcolor{blue}{1b}   & \textcolor{blue}{2b1}   & \textcolor{blue}{1a}   & \textcolor{green}{9b}   & \textcolor{green}{12a1} \\
Italian  & \textcolor{green}{12b1}& \textcolor{blue}{1b}    & \textcolor{green}{9b}  & \textcolor{green}{5b1}  & \textcolor{red}{6a}     \\
Russian  & \textcolor{blue}{1b}   & \textcolor{green}{12b1} & \textcolor{green}{9b}  & \textcolor{red}{3b}     & \textcolor{green}{8b2}  \\
Japanese & \textcolor{green}{12b2}& \textcolor{blue}{2b1}   & \textcolor{blue}{1b}   & \textcolor{green}{9b}   & \textcolor{blue}{1a}    \\
\bottomrule
\end{tabular}
\end{table}

\begin{table}[t]
\caption{Top-5 worst answered templates by language across all models. Templates for questions about the following event are in \textcolor{blue}{blue}, templates for questions about preceding events in \textcolor{red}{red}, and templates for questions about simultaneous events in \textcolor{green}{green}.}
\label{tab:worst5_templates}
\centering
\small
\begin{tabular}{lccccc}
\toprule
\textbf{} & \textbf{1} & \textbf{2} & \textbf{3} & \textbf{4} & \textbf{5} \\
\midrule
English  & \textcolor{green}{12b1} & \textcolor{red}{6b} & \textcolor{green}{12b2} & \textcolor{blue}{10a} & \textcolor{blue}{11a} \\
Italian  & \textcolor{red}{3b} & \textcolor{red}{4b} & \textcolor{green}{12b2} & \textcolor{red}{7b} & \textcolor{blue}{2b2} \\
Russian  & \textcolor{red}{7b} & \textcolor{blue}{10b} & \textcolor{red}{6b} & \textcolor{red}{4b} & \textcolor{blue}{2b2} \\
Japanese & \textcolor{red}{7b} & \textcolor{red}{4b} & \textcolor{red}{6b} & \textcolor{blue}{10a} & \textcolor{blue}{11a} \\
\bottomrule
\end{tabular}
\end{table}

\begin{table}[t]
\caption{Top-5 best answered templates by model across all languages. Templates for questions about the following event are in \textcolor{blue}{blue}, templates for questions about preceding events in \textcolor{red}{red}, and templates for questions about simultaneous events in \textcolor{green}{green}.}
\label{tab:top5_model_templates}
\centering
\small
\begin{tabular}{lccccc}
\toprule
\textbf{} & \textbf{1} & \textbf{2} & \textbf{3} & \textbf{4} & \textbf{5} \\
\midrule
GPT-4o                 & \textcolor{green}{9b}   & \textcolor{blue}{2b2}   & \textcolor{green}{8b2}   & \textcolor{blue}{2b1}   & \textcolor{green}{5b1} \\
Gemini-2.0-flash-lite   & \textcolor{blue}{1b}    & \textcolor{green}{9b}   & \textcolor{blue}{1a}     & \textcolor{red}{6a}     & \textcolor{green}{8b1} \\
Qwen2-VL-7B-Instruct    & \textcolor{green}{12b1} & \textcolor{green}{12a1} & \textcolor{blue}{2b1}    & \textcolor{blue}{1a}    & \textcolor{green}{12a2} \\
MiniCPM-V-2\_6          & \textcolor{green}{12b1} & \textcolor{blue}{1b}    & \textcolor{blue}{2b1}    & \textcolor{green}{9b}   & \textcolor{green}{5b2} \\
InternVL2-8B            & \textcolor{green}{12b1} & \textcolor{green}{12b2} & \textcolor{red}{3b}      & \textcolor{blue}{1b}    & \textcolor{blue}{10b} \\
\bottomrule
\end{tabular}
\end{table}

\begin{table}[t]
\caption{Top-5 worst answered templates by model across all languages. Templates for questions about the following event are in \textcolor{blue}{blue}, templates for questions about preceding events in \textcolor{red}{red}, and templates for questions about simultaneous events in \textcolor{green}{green}.}
\label{tab:worst5_model_templates}
\centering
\small
\begin{tabular}{lccccc}
\toprule
\textbf{} & \textbf{1} & \textbf{2} & \textbf{3} & \textbf{4} & \textbf{5} \\
\midrule
GPT-4o                & \textcolor{green}{12b1} & \textcolor{red}{4b}   & \textcolor{blue}{10a}   & \textcolor{red}{7b}    & \textcolor{green}{5a1} \\
Gemini-2.0-flash-lite & \textcolor{red}{7b}     & \textcolor{red}{3b}   & \textcolor{green}{12b1} & \textcolor{red}{6b}    & \textcolor{blue}{10a}  \\
Qwen2-VL-7B-Instruct  & \textcolor{blue}{2b2}   & \textcolor{red}{7b}   & \textcolor{blue}{10b}   & \textcolor{blue}{11a}  & \textcolor{blue}{11b}  \\
MiniCPM-V-2\_6        & \textcolor{red}{6b}     & \textcolor{blue}{10b} & \textcolor{red}{7b}     & \textcolor{red}{4b}    & \textcolor{red}{3b}    \\
InternVL2-8B          & \textcolor{red}{4b}     & \textcolor{red}{7b}   & \textcolor{green}{9a}   & \textcolor{blue}{2b2}  & \textcolor{blue}{11b}  \\
\bottomrule
\end{tabular}
\end{table}

